\documentclass[10pt]{article}
\usepackage[latin9]{inputenc}
\usepackage{listings}
\usepackage{amsmath}
\usepackage{amssymb}
\usepackage{graphicx}
\usepackage{subscript}
\usepackage{nameref}

\makeatletter

\newcommand{\noun}[1]{\textsc{#1}}

\newenvironment{lyxlist}[1]
{\begin{list}{}
{\settowidth{\labelwidth}{#1}
 \setlength{\leftmargin}{\labelwidth}
 \addtolength{\leftmargin}{\labelsep}
 }}
{\end{list}}

\usepackage{float}\usepackage{scalefnt}\usepackage{ifthen}\usepackage{colortbl}\usepackage{multirow}\usepackage{array}\usepackage{listings}

\let\textquotedbl="

\makeatother

\begin{document}

\title{Learning Language from a Large (Unannotated) Corpus }

\author{\textit{Linas Vepstas and Ben Goertzel}}
\maketitle
\begin{abstract}
A novel approach to the fully automated, unsupervised extraction of
dependency grammars and associated syntax-to-semantic-relationship
mappings from large text corpora is described. The suggested approach
builds on the authors' prior work with the Link Grammar, RelEx and
OpenCog systems, as well as on a number of prior papers and approaches
from the statistical language learning literature. If successful,
this approach would enable the mining of all the information needed
to power a natural language comprehension and generation system, directly
from a large, unannotated corpus. 
\end{abstract}

\section{Introduction}

Currently, the two primary methods of supplying natural language processing
systems with \textquotedbl{}content\textquotedbl{} regarding specific
languages are:
\begin{enumerate}
\item Explicit human-coded linguistic rules,
\item Supervised machine learning from human-annotated corpora.
\end{enumerate}
\noindent Neither of these approaches is fully satisfactory, because
both rely on substantial formalized coding by expert humans. Natural
language is sufficiently complex and diverse that it eludes full formalization,
either in the form of hand-coded rules, or in the form of annotation
of corpora. Even if, in principle, some sufficiently large hand-coded
rule-set or annotated corpus was enough to supply an NLP system with
linguistic content, it leaves open the question of operating with
the higher, more abstract structures that are the outcome of parsing:
rule-sets do not address the issue of semantic content. Thus, in practice,
these traditional approaches are unlikely to yield full success.

The goal here is to explore the alternative: the induction of grammar
and semantics by means of unsupervised learning algorithms. Two approaches
that have been discussed and attempted, to some extent, by the scientific
community, are:
\begin{enumerate}
\item Machine learning from large, unannotated text corpora 
\item Machine learning from (unannotated) data regarding spoken or textual
language in non-linguistic contexts (\emph{e.g.} texts together with
pictures, or spoken language together with video and ambient audio). 
\end{enumerate}
\noindent Interesting ideas have been developed in both of these directions,
but, so far, results have fallen far short of those obtained via the
first two approaches.

The review of \cite{Klein2004} provides a summary of the state of
the art in automatic grammar induction (the third alternative listed
above), as it stood a decade ago: it addresses a number of linguistic
issues and difficulties that arise in actual implementations of algorithms.
It is also notable in that it builds a bridge between phrase-structure
grammars and dependency grammars, essentially pointing out that these
are more or less equivalent, and that, in fact, significant progress
can be achieved by taking on both points of view at once. Grammar
induction has progressed somewhat since this review was written, and
we will mention some of the more recent work below; but yet, it is
fair to say that there has been no truly dramatic progress in this
direction.

Here, we describe a novel approach to achieving the third alternative:
automated grammar induction by machine learning of linguistic content
from a large, unannotated text corpus. The methods described may also
be useful for the fourth alternative (incorporation of extralinguistic
data in the learning system's inputs); and could make use of content
created using hand-coded rules or machine learning from annotated
corpora. However, our focus will be on learning linguistic content
from a large, unannotated text corpus.

While the overall approach presented here is novel, the ideas are
extensions and generalizations of the prior work of multiple authors,
which will be referenced and in some cases discussed below. We believe
the body of ideas needed to enable unsupervised learning of language
from large corpora has been gradually emerging during the last decade.
The approach given here has unique aspects, but also many aspects
have already been validated by the work of others.

For sake of simplicity, we will deal here only with learning from
written text. We believe that conceptually similar methods can be
applied to spoken language as well, but that this brings extra complexities
that we will avoid for the purposes of the present document. (In short:
below, we represent syntactic and semantic learning as separate but
similarly structured and closely coupled learning processes. To handle
speech input thoroughly, we would suggest phonological learning as
another separate, similarly structured and closely coupled learning
process.)

Finally, we stress that the algorithms presented here are intended
to be used in conjunction with a large corpus, and a large amount
of processing power. Without a very large corpus, some of the feedback
required for the learning process described would be unlikely to happen
(\emph{e.g.} the ability of syntactic and semantic learning to guide
each other). We have not yet sought to estimate exactly \textit{how}
large a corpus would be required, but our informal estimate is that
Wikipedia might or might not be large enough, and the Web is certainly
more than enough.

\section{Algorithmic Overview}

The rest of this paper is devoted to fleshing out, providing detail,
and mounting a theoretical defense of a rather simple, basic algorithm.
Rather than getting lost in the details, it is important to keep a
general notion of the algorithm in mind at all times. Thus, a crude
sketch follows.

The algorithm is as follows: Step A) Define words to be 'things';
Step B) Look for correlations between things; Step C) Cluster similar
things together into classes; Step D) Define a new set of things as
the clusters obtained from the last step and Step E) return to step
B. By correlation, it will almost always be meant 'mutual information'
or 'mutual entropy'. This is a number capturing the strength of a
relationship between 'things'. The 'relationship' between things will
be, in general, a graph or hypergraph. However, very early in the
iteration of the algorithm, it will be very simple: if 'things' are
words, then the 'relationship' is pairs of words, and one starts by
measuring the mutual information of pairs of words.

The classification of step C is to be accomplished primarily using
entropy maximization/minimization principles. To illustrate with pairs
of words: consider words A, B and W. Then, word A and word B should
be grouped together into a cluster C, if, for any word W, the total
entropy of \emph{pair}(word W, class C) + \emph{member}(A in C) +
\emph{member}(B in C) is less than the total entropy of \emph{pair}(word
W, word A) + \emph{pair}(word W, word B). If this inequality holds,
then the cluster C should be formed; if not, then there is no advantage
to having such a class, and it should be dissolved. (This example
should not be taken literally, as, even for word pairings, the actual
relationships that must be considered are more complex than this.
Detailed inequalities for clustering are presented in an appendix).

The penultimate step D makes considerable demands on technology. In
order to use 'things' observed in the environment, one must be able
to recognize those things: and so, one needs to have a pattern recognizer.
However, patterns do not sit all by themselves, but, in fact, they
interconnect, and so one must have a way assembling them so that they
match the observed input. This is the 'parser' of natural language
processing. Here, in the abstract context of 'patterns' of 'things',
its probably best to think of each pattern as a puzzle-piece. A collection
of puzzle pieces must then be assembled into a complete, final picture
that more or less matches the observed input. This is essentially
a ``constraint satisfaction problem'', which should conjure up the
kinds of algorithms required, as well as the difficulties one may
have in recognizing patterns in an input. For the general case, the
best and most well-known algorithm for solving constraints is the
DPLL algorithm\cite{WP-DPLL}, on which most SAT solvers\cite{WP-SAT}
and other systems are based. For language learning, though, it does
not seem that a jump to SAT is immediately required. Many patterns
will be relatively simple, and have simple inter-dependencies, for
which more basic algorithms should suffice. These include the backwards-forwards
or the Viterbi algorithms, which should be fast, effective and sufficient.
So, for example, in parsing a sentence, as each new word is 'heard',
a set of different relationships with the surrounding words may be
contemplated. After enough words have been heard in a row (say 5 or
7 or 10), their inter-relationship should become clear, and one can
move on.%
\footnote{This is anchored on the psycho-linguistic observation that almost
all dependencies are short, and rarely extend past the the first few
nearest neighbors.\cite{Gibson1998,Temper2007,Liu2008,Ferrer2006}%
} This is essentially the Viterbi algorithm, which discards the unlikely
combinations, keeping only the most likely candidate(s). A global,
Boolean-SAT-style optimization is not required: what we hear now really
should not affect the parse of a sentence we heard several minutes
ago. However, this changes once the patterns become complex enough.
So, for example, consider a set of large patterns, encoding meaning,
that span multiple sentences or even paragraphs. These large patterns
may not fit together in any simple way, and may require a good amount
of wrestling to assemble together in a coherent, consistent fashion.
This is where the full force of a strong constraint-satisfaction solver
would be required.

Implicit in step D is also a layering or recursion of pattern complexity,
and the application of 'deep learning' principles. The groupings of
the previous step have a tendency to reduce the total number of rules
or relationships needed to describe the corpus (entropy minimization);
however, the complexity of the rules tends to increase. It is the
sum of the (logarithm of) the number of rules, and the (logarithmic)
complexity that provides the proper metric: an ``Occam's razor''
to discover the least-complex and smallest set of rules possible.
However, as confidence in a set of rules grows, and uncertainty diminishes,
exceptions become more apparent. Such exceptions encode semantic information.
So, for example, commonality between the phrases ``the book has ...''
and ``the dog has ...'' suggests that ``book'' and ``dog'' can
be grouped into a class ``noun''. The absence of phrases such as
``the book barked at the squirrel'' suggests that perhaps books
and dogs differ, and can be sub-classified as inanimate and animate
objects. This re-classification has the odd effect of strengthening
the original conception of ``noun'', by forcing out words that were
incorrectly classified as nouns in the first place. This can be viewed
as a form of ``deep learning'' at work: a higher, more abstract
layer can serve to refine the correctness of a shallower, more concrete
layer. Implicit in the algorithm is not just an inter-dependency of
rules, but also a layering or hierarchy.

This, then, is the basic outline of the algorithm that is being proposed
here. It should be clear, from this description, that it is capable
of levering it's way up from a lack of structure to a collection of
complex patterns, and doing so without any 'training data', in an
unsupervised fashion. The remainder of this proposal is then devoted
to justifying why this might be the right approach, as well as fleshing
out in greater detail in how all this might work.

\section{Assumed Linguistic Infrastructure}

While the approach outlined here aims to learn the linguistic content
of a language from textual data, it does not aim to learn the \textit{idea}
of language. Implicitly, we assume a model in which a learning system
begins with a basic \textquotedbl{}linguistic infrastructure\textquotedbl{}
indicating the various parts of a natural language and how they generally
interrelate; and it then learns the linguistic content characterizing
a particular language. In principle, it would also be possible to
have an AI system to learn the very concept of a language and build
its own linguistic infrastructure. However, that is not the problem
we address here; and we suspect such an approach would require drastically
more computational resources.

The basic linguistic infrastructure assumed here includes:

\begin{figure}
\caption{Dependency and Phrase-Structure Parses\label{fig:Dependency-and-Phrase-Structure}}

\begin{centering}
\includegraphics[width=12cm]{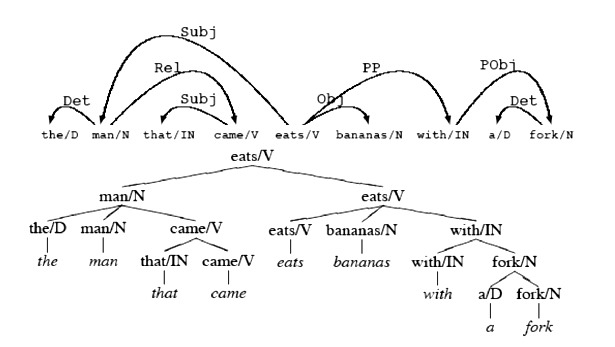} 
\par\end{centering}

A comparison of dependency and phrase-structure parses, above and
below. In general, one can be converted to the other (algorithmically);
dependency parses tend to be easier understand and verify. In the
dependency parse, an arrow points from the controlling word or head
word to the dependent word. (Somewhat confusingly, the head of the
arrow points at the dependent word; this means the tail of the arrow
is attached to the head word). (Image taken from G. Schneider, ``Learning
to Disambiguate Syntactic Relations'' Linguistik online 17, 5/03)\\
 \rule[0.5ex]{1\columnwidth}{1pt} 
\end{figure}

\begin{figure}
\begin{centering}
\caption{Link Grammar Connectors\label{fig:Link-Grammar-Connectors}}

\par\end{centering}

\begin{centering}
\includegraphics[width=0.9\columnwidth]{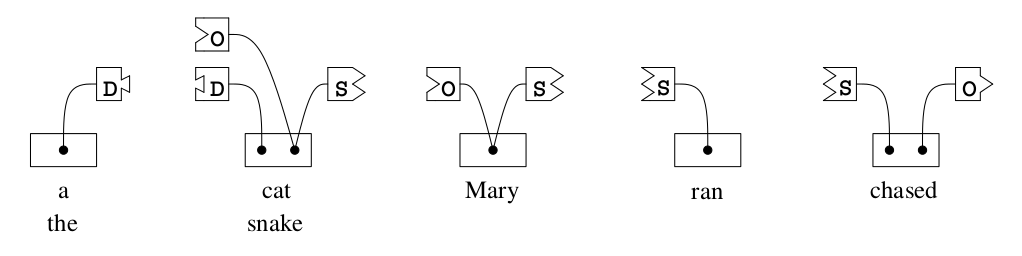}
\par\end{centering}

An illustration of Link Grammar connectors and disjuncts. The connectors
are the jigsaw-puzzle-shaped pieces; connectors are allowed to connect
only when the tabs fit together. A disjunct is the entire (ordered)
set of connectors for a word. As lexical entries appearing in a dictionary,
the above would be written as 
\begin{lstlisting}
		a the: D+;
		cat snake: D- & (S+ or O-);
		Mary: O- or S+;
		ran: S-;
		chased S- & O+;
\end{lstlisting}
Note that although the symbols \texttt{``\&''} and \texttt{``or''}
are used to write down disjuncts, these are \textbf{\emph{not}} Boolean
operators, and do \textbf{\emph{not}} form a Boolean algebra. They
do form a non-symmetric compact closed monoidal algebra. The diagram
below illustrates puzzle pieces, assembled to form a parse: 

\begin{centering}
\includegraphics[width=0.65\columnwidth]{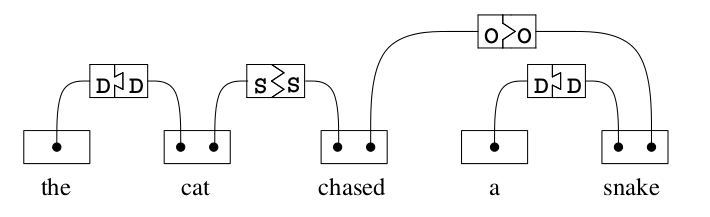} 
\par\end{centering}

The comparable ASCII-graphics parse, from a recent version of the
parser, is: 

\begin{centering}
\begin{lstlisting}
		               +----Os----+  
		 +-Ds-+---Ss---+     +-Ds-+
		 |    |        |     |    |
		the cat.n chased.v-d a snake.n  
\end{lstlisting}

\par\end{centering}

The additional lower-case '\texttt{s}' shown here (\emph{e.g.} \texttt{Ds})
indicates that the link connects to a singular (not plural) noun.
The words have also been decorated with parts of speech. (Images taken
from \cite{Sleator1991}.)\\
 \rule[0.5ex]{1\columnwidth}{1pt} 
\end{figure}

\begin{itemize}
\item A formalism for expressing grammatical (dependency) rules is assumed.

\begin{itemize}
\item The ideas given here are not tied to any specific grammatical formalism,
but we find it convenient to make use of a formalism in the style
of dependency grammar\cite{Tesn1959}. Taking a mathematical perspective,
different grammar formalisms can be translated into one-another, using
relatively simple rules and algorithms\cite{Klein2004}. The primary
difference between them is more a matter of taste, perceived linguistic
'naturalness', adaptability, and choice of parser algorithm. In particular,
categorial grammars\cite{Kart2013} can be converted into link grammars
in a straight-forward way, and \emph{vice versa}, but link grammars
provide a more compact dictionary. Link grammars\cite{Sleator1991,Sleator1993}
are a type of dependency grammar; these, in turn, can be converted
to and from phrase-structure grammars. We believe that dependency
grammars provide a more simple and natural description of linguistic
phenomena. We also believe that dependency grammars have a more natural
fit with maximum-entropy ideas, where a dependency relationship can
be literally interpreted as the mutual information between word-pairs\cite{Yuret1998}.
Dependency grammars also work well with Markov models; dependency
parsers can be implemented as Viterbi decoders. Figure \ref{fig:Dependency-and-Phrase-Structure}
illustrates two different formalisms. 
\item The discussion below assumes the use of a formalism similar that of
Link Grammar. In this theory, each word is associated with a set of
'connector disjuncts', each connector disjunct controlling the possible
linkages that the word may take part in. A disjunct can be thought
of as a jig-saw puzzle-piece; valid syntactic word orders are those
for which the puzzle-pieces can be validly connected. A single connector
can be thought of as a single tab on a puzzle-piece (shown in figure
\ref{fig:Link-Grammar-Connectors}). Connectors are thus 'types' $X$
with a + or - sign indicating that they connect to the left or right.
For example, a typical verb disjunct might be $S-\&\, O+\,$ indicating
that a subject (a noun) is expected on the left, and an object (also
a noun) is expected on the right. 
\item Some of the discussion below assumes aspects of (Dick Hudson's) Word
Grammar\cite{Hud84,Hud07}. This theory (implicitly) uses connectors
similar to those of Link Grammar, but allows each connector to be
marked as the head of a link or not. A link then becomes an arrow
from a head word to the dependent word. 
\item Each word is associated with a ``lexical entry''; in Link Grammar,
this is the set of connector disjuncts for that word. It is usually
the case that many words share a common lexical entry; for example,
most common nouns are syntactically similar enough that they can all
be grouped under a single lexical entry. Conversely, a single word
is allowed to have multiple lexical entries; so, for example, ``saw'',
the noun, will have a different lexical entry from ``saw'', the
past tense of the verb ``to see''. That is, lexical entries can
loosely correspond to traditional dictionary entries. Whether or not
a word has multiple lexical entries is a matter of convenience, rather
than a fundamental aspect. Curiously, a single Link Grammar connector
disjunct can be viewed as a very fine-grained part-of-speech. For
example, $S-\&\, O+\,$ is the disjunct used for transitive verbs
(verbs that take an object), while $S-\&\, O+\&\, On+\,$ is the disjunct
for ditransitive verbs (verbs that take two objects: a direct and
indirect object). These fine-grained parts-of-speech correlate reasonably
well with word senses (e.g. taken from WordNet) and can thus serve
as a rough (and very rapid!) suggestion for word-sense disambiguation.
In this way, disjuncts are a stepping stone to the semantic meaning
of a word. 
\end{itemize}
\item A parser, for extracting syntactic structure from sentences, is assumed.
What's more, it is assumed that the parser is capable of using dynamic
criteria, such as semantic relationships, to guide parsing.

\begin{itemize}
\item The statement here is about the type of functionality needed from
the parsing component. Traditional parsers presume a static, fixed
lexis, and do not provide any mechanism by which parse ranking can
be adjusted or steered, on-the-fly, by mechanisms outside of the domain
of the parser itself (``deep learning''). Yet, such external influence
seems centrally important to a realistic system.
\item A paradigmatic example of such a parser is the ``Viterbi Link Parser'',
currently under development for use with the Link Grammar. This parser
is currently operational in a simple form. The name refers to its
use of the general ideas of the Viterbi algorithm. This algorithm
seems biologically plausible, in that it applies only a local analysis
of sentence structure, of limited scope, as opposed to a global optimization,
thus roughly emulating the process of human listening.%
\footnote{Again, this is anchored on the observation that almost all dependencies
are short, and rarely extend past the the first few nearest neighbors.\cite{Gibson1998,Temper2007,Liu2008,Ferrer2006}%
} The current set of legal parses of a sentence is pruned incrementally
and probabilistically, based on flexible criteria. Although the core
criteria are meant to be the traditional grammatical dependency rules
taken from the lexis, they need not be limited to this. Thus, criteria
that can sway parse likelihood potentially include the semantic classes
and roles extracted from a partial parse obtained at a given point
in time; such semantic relationships are typically not present in
a traditional syntactic lexis. Dynamic likelihood criteria also allows
for parsing to be guided by inter-sentence relationships, such as
pronoun resolution, to disambiguate otherwise ambiguous sentences.
\item Inherent in parsing is assigning a likelihood or probability to a
given parse. This probability is assembled from several sources, including
an inherent strength of relationships (some disjuncts are inherently
more appropriate than others), structural constraints (long range
relationships and link-crossings are disfavored) as well as a combinatorial
entropy of possible choices (``\emph{Sri Lanka}'' is a set phrase).
\end{itemize}
\item A formalism for expressing semantic relationships is assumed.

\begin{itemize}
\item A semantic relationship generalizes the notion of a lexical entry
to allow for changes of word order, paraphrasing, tense, number, the
presence or absence of modifiers, \emph{etc.} An example of such a
relationship would be \emph{eat(X, Y)} -- indicating the eating of
some entity \emph{Y} by some entity \emph{X. }This abstracts into
common form several different syntactic expressions: ``\emph{Ben
ate a cookie}'', ``\emph{A cookie will be eaten by Ben}'', ``\emph{Ben
sat, eating cookies}''. 
\item Nothing particularly special is assumed here regarding semantic relationships,
beyond a basic predicate-argument structure\cite{WP-Predicate,WP-Argument}.
It \emph{is} assumed that predicates can have arguments that are other
predicates, and not just atomic terms; this has an explicit impact
on how predicates and arguments are represented. A ``semantic representation''
of a sentence is a network of arrows (defining predicates and arguments),
each arrow or a small subset of arrows defining a ``semantic relationship''.
However, the beginning or end of an arrow is not necessarily a single
node, but may land on a subgraph. Because arrows may point from to
to subgraphs, the resulting structure itself is no longer a graph
in the proper sense, but a hypergraph.
\item Type constraints seem reasonable, but its not clear if these must
be made explicit, or if they are the implicit result of learning.
Thus, \emph{eat(X, Y)} requires that \emph{X} and \emph{Y} both be
entities, and not, for example, actions or prepositions. 
\end{itemize}
\item A formalism for expressing topics, themes and discourse structure
is assumed.

\begin{itemize}
\item Topics, themes and general discourse structure are concepts that are
probed at the more abstract levels of linguistic structure. A particularly
appealing form of these, at least from the algorithmic, computer-science
perspective, is Mel'\v{c}uk's Meaning-Text Theory (MTT).\cite{Mel'cuk1987,Kahane2003}
The reason for this appeal is that the theory is based on a sequence
of transformations on graphs (some of these apparently being 'natural
transformations' in the category-theoretic sense), and thus amenable
to precise formalization and an algorithmic treatment. A very short
review of MTT is provided in an appendix.
\item The formalism, insofar as it is ultimately graphical in nature, does
not really extend much past that required for describing syntactic
disjuncts or semantic relationships. Rather, the point here is that
one can have not just 2-point relationships, such as \emph{eat(X,
Y)}, but more generally, \emph{n}-point relations which themselves
may have some internal structure. Thus, one may write $r(x_{1},\cdots,x_{n})$
for an n-point relation (or constraint) between objects, but also,
specific relations may themselves have a hierarchical structure $r(x_{1},\cdots,r_{k}(y_{1},\cdots,y_{m}),\cdots,x_{n})$
encapsulating a required graphical sub-structure. 
\end{itemize}
\end{itemize}
The above summarizes the basic software components and theoretical
linguistic infrastructure that is presumed, entering into the exercise.
To summarize the above from a computational or algorithmic viewpoint,
it is presumed that linguistic relationships and dependencies can
be captured in the form of relations $r(x_{1},\cdots,x_{n})$, between
items $x_{1},\cdots,x_{n}$, together with a parser (or pattern-matcher
or constraint-solver) that can, given a lexis of relations, and an
input, can find a set of relations that, like puzzle-pieces, assemble
to reproduce the input. 

To be concrete, a Link Grammar connector $S-$ can be more abstractly
written as a relation $r(w_{l},w_{\bullet},t=S)$ denoting that the
current word $w_{\bullet}$ must attach to a word $w_{l}$ to the
left, using a connector type $S$. The Link Grammar constraint $\&$
specifies that several connectors must be present and connected, and
might be written as a relation $r(c_{1},c_{2})$ where each $c_{k}$
is a connector. The hierarchical nesting arises from the need to represent
a Link Grammar disjunct. So, a transitive verb, normally expressed
as $S-\&\, O+\,$, might now be written as $r(c(w_{l},w_{\bullet},t=S),\: c(w_{\bullet},w_{r},t=O))$
indicating that the verb $w_{\bullet}$ must have a subject $w_{l}$
on the left and an object $w_{r}$ on the right. However, in the abstract,
such a relationship is not fundamentally different from one such as
$r(X,Y,s=eat)$ for the semantic relation \emph{eat(X,Y)}, or the
relation $r(f=magnitude,\ noun=rain,\ modifier=torrential)$ for the
lexical function that associates magnitude modifiers with the object
being modified.

The role of the parser is to take input at one abstraction level,
and generate an output at the next abstraction level. At the lowest
level, parsing consists of taking an ordered string of words, and
generating a set of dependency relationships. At the next level, parsing
consists of taking a set of dependency relationships, and extracting
semantic relationships. In either case, the only allowed parses are
those that fulfill the constraints imposed by the (currently known)
relationships. These constraints include structural relations (\emph{e.g.}
being to the left of) as well as type relations (\emph{e.g.} being
a noun).

The point here is that, although we listed the assumed infrastructure
in linguistic terminology, the actual required infrastructure is not
inherently linguistic: rather, it is a system of constraints, each
enforced with some probability or strength, used to analyze an input,
and discover the (graphical, typed relationship) structure within
it. We believe that this generic infrastructure can be applied to
domains outside of linguistics, but will not dwell further on this
point here.

\section{Linguistic Content To Be Learned}

Given the above linguistic infrastructure, what remains for a language
learning system to learn is the \textit{linguistic content} that characterizes
a particular language. Specifically, given the assumed framework,
key items to be learned are listed below. These are listed roughly
in order of sophistication and complexity, with the earlier elements
being easier to learn, and being learnt earlier, than the later items.

Although the previous section concluded with an abstract view of the
infrastructure, of being merely a collection of structural relationships,
we revert back to a set of concrete tasks to be achieved. Thus we
have the following:
\begin{itemize}
\item A list of 'link types' that will be used to form 'disjuncts' must
be learned.

\begin{itemize}
\item An example of a link type is the 'subject' link $S$. This link typically
connects the subject of a sentence to the head verb. Given the normal
English subject-verb word order, nouns will typically have an $S+$connector,
indicating that an $S$ link may be formed only when the noun appears
to the left of a word bearing an $S-$ connector. Likewise, verbs
will typically be associated with $S-$ connectors. The current Link
Grammar contains roughly one hundred different link-types, with additional
optional subtypes that are used to further constrain syntactic structure.
This number of different link types seems required simply because
there are many relationships between words: there is not just a subject-verb
or verb-object relationship, but also rather fine distinctions, such
as those needed to form grammatical time, date, money, and measurement
expressions, punctuation use, including street-addresses, cardinal
and ordinal relationships, proper (given) names, titles and suffixes,
and other highly constrained grammatical constructions. This is in
addition to the usual linguistic territory of needing to indicate
dependent clauses, comparatives, subject-verb inversion, and so on.
It is expected that a comparable number of link types will need to
be learned. 
\item Some link types are rather strict, such as those that connect verb
subjects and objects, while other types are considerably more ambiguous,
such as those involving prepositions. This reflects the structure
of English, where subject-verb-object order is fairly rigorously enforced,
but the ordering and use of prepositions is considerably looser. When
considering the looser cases, it becomes clear that there is no single,
inherent 'right answer' for the creation and assignment of link types,
and that several different, yet linguistically plausible linkage assignments
may be made. 
\item The definition of a good link-type is one that leads the parser --
applied across the whole corpus -- to allow parsing to be successful
for almost all sentences, and yet not to be so broad as to enable
parsing of word-salads. Significant pressure must be applied to prevent
excess proliferation of link types, yet no so much as to over-simplify
things, and provide valid parses for unobserved, ungrammatical sentences. 
\end{itemize}
\item Lexical entries for different words must be learned.

\begin{itemize}
\item Typically, multiple connectors are needed to define how a word can
link syntactically to others. Thus, for example, many verbs have the
disjunct $S-\&\, O+\,$ indicating that they need a subject noun to
the left, and an object to the right. All words have at least a handful
of valid disjuncts that they can be used with, and sometimes hundreds
or even more. Thus, a ``lexical entry'' must be learned for each
word, the lexical entry being a set of disjuncts that can be used
with that word. 
\item Many words are syntactically similar; most common nouns can share
a single lexical entry. Yet, there are many exceptions. Thus, during
learning, there is a back-and forth process of grouping and ungrouping
words; clustering them so that they share lexical entries, but also
splitting apart clusters when its realized that some words behave
differently. Thus for example, the words ``sing'' and ``apologize''
are both verbs, and thus share some linguistic structure, but one
cannot say ``I apologized a song to Vicky'' because apologize is
not a ditransitive verb; if these two verbs were initially grouped
together into a common lexical entry, they must later be split apart. 
\item The definition of a good lexical entry is much the same as that for
a good link type: observed sentences must be parsable; random sentences
mostly must not be, and excessive proliferation and complexity must
be prevented. 
\end{itemize}
\end{itemize}
We pause here to observe that the distinction between purely syntactic
relations, and those with semantic overtones, can be blurry. Basic
semantic content can be derived from exceptions to over-generalized
syntactic rules, or from a narrowing of the applicability of such
rules to finer classes. Thus, for example, \emph{``the book chased
a squirrel}'' is not likely to be observed during a scan of a large
corpus. This can be dealt with at the lexical level: it is a mistake
to place \emph{``book}'' into a class of ``nouns''; rather, it
belongs to a class of ``inanimate nouns''. Likewise, ``\emph{chase}''
is not merely some transitive verb, but a verb that can only attach
to animate subjects. With sufficient parsimony pressure, it seems
reasonable that such finer semantic distinctions could be learned
at what naively appears to be a purely syntactic level. The extent
to which this might take place depends heavily on the metaphoric content
of the input corpus: a literary review might contain a sentence ``\emph{the
book chased an absurd premise}'' suggesting naively that books perhaps
are animate. Again, semantic content can appear as exceptions to generalized
rules.
\begin{itemize}
\item Semantic relationships must be learned.

\begin{itemize}
\item The semantic relationship \emph{eat(X,Y)} is prototypical. Foundationally,
such a semantic relationship may be represented as a set whose elements
consist of syntactico-semantic subgraphs. For the relation \emph{eat(X,Y)},
a subgraph may be as simple as a single (syntactic) disjunct $S-\&\, O+\,$
for the normal word order ``\emph{Ben ate a cookie}'', but it may
also be a more complex set needed to represent the inverted word order
in \emph{``a cookie was eaten by Ben}''. 
\item The task here is then to learn synonymous re-phrasings: not just sets
of words that are synonyms, but phrases. These need not be centered
on a verb, so that ``\emph{Wyoming borders on Colorado}'' is synonymous
to to ``\emph{Colorado is a neighbor of Wyoming}'', and both are
captured by a prepositional relation ``\emph{next\_to(Colorado, Wyoming)}''.
Such re-phrasings are at a different abstraction level from the syntactic
parse level.
\item The set of all of these different subgraphs defines the semantic relationship.
The subgraphs themselves may be syntactic (as in the examples above),
or they may be other semantic relationships, or a mixture thereof. 
\item Not all re-phrasings are semantically equivalent. ``\emph{Mr. Smith
is late}'' has a rather different meaning from ``\emph{The late
Mr. Smith.}'' 
\item In general, place-holders like X and Y may be words or category labels.
In early stages of learning, it is expected that \emph{X} and \emph{Y}
are each just sets of words. At some point, though, it should become
clear that these sets are not specific to this one relationship, but
can appropriately take part in many relationships. In the above example,
X and Y must be entities (physical objects), and, as such, can participate
in (most) any other relationships where entities are called for. More
narrowly, \emph{X} is presumably a person or animal, while \emph{Y}
is a foodstuff. Furthermore, as entities, it might be inferred when
these refer to the same physical object (see the section 'reference
resolution' below). 
\item Categories can be understood as sets of synonyms, including hyponyms
(thus, ``\emph{grub}'' is a synonym for ``\emph{food}'', while
``\emph{cookie''} is a hyponym. 
\end{itemize}
\item Idioms and set phrases must be learned.

\begin{itemize}
\item English has a large number of idiomatic expressions whose meanings
cannot be inferred from the constituent words (such as ``\emph{to
pull one's leg}''). In this way, idioms present a challenge: their
sometimes complex syntactic constructions belie their often simpler
semantic content. On the other hand, idioms have a very rigid word-choice
and word order, and are highly invariant. Set phrases take a middle
ground: word-choice is not quite as fixed as for idioms, but, none-the-less,
there is a conventional word order that is usually employed. Note
that the manually-constructed Link Grammar dictionaries contain thousands
of lexical entries for idiomatic constructions. In essence, these
are multi-word constructions that are treated as if they were a single
word. 
\end{itemize}
\end{itemize}
Each of the above tasks have already been accomplished and described
in the literature; for example, automated learning of synonymous words
and phrases has been described by Lin\cite{Lin2001} and Poon \& Domingos\cite{Poon2009}.
However, Lin and Poon \& Domingos each assume the pre-existence of
a syntactic dependency parser, rather than starting from ground zero.
The authors are not aware of any attempts to learn all of these, together,
in one go, rather than presuming the pre-existence of dependent layers.

Furthermore, no previous work has attempted to attack language learning
in a fully abstract structural setting, although in some sense, work
within the framework of Markov Logic Networks (MLN), such as that
of Poon \& Domingos, comes close. Taken at an abstract level, MLN
combines several distinct components: the notion of a network or graph
(and thus similar to the notion of a Bayesian network; the term ``Markovian''
referring to a certain independence assumption), the idea that a network
can express first-order logic (or, more generally, the internal language
of category theory), and finally, that unknowns must be distributed
uniformly in probability space (by applying a maximum entropy principle).
One difficulty with MLN is common to all maximum-entropy (ME) approaches,
and that is that solving for the Lagrange variables of the ME equations
is an NP-hard problem: the potential function can be riddled with
local maxima; hill-climbing may be slow and converge to an inappropriate
solution. Thus, while in principle, it is useful to express fealty
to the notion of evenly distributing unknowns, in practice, the algorithms
can be slow to converge. As a result of this, another commonly successful
approach is that of Bayesian networks (such as Hidden Markov Models).
Here, probabilities are assigned by a different algorithm (following
from a naive application of Bayes theorem) that is usually far faster.
However, the application of naive Bayes almost immediately breaks
down due to the need for independence assumptions. So, for example,
while ``\emph{Sri Lanka}'' is, formally, two distinct words, these
cannot be treated independently of one another: it is nearly impossible,
in English, to use the one word without the other, and so, from the
combinatorial viewpoint, these must really count as only one word.
Bayesian approaches typically have difficulty with counting. Thus,
neither Bayesian nor MLN approaches are entirely satisfactory; a different
algorithmic approach is sought, while keeping in place the fundamental
concept of a network of relations.

Thus, to return to the abstract setting outlined at the end of the
previous section: we wish to learn a lexis of relations $r(x_{1},\cdots,x_{n})$.
Treated as relations, these can be viewed as forming a ``network''.
Insofar as as they are constraints with variables, they can be viewed
as a form of ``logic'', although the constraints are not those of
first-order logic (as is amply clear from the Link Grammar operators
\& and ``or'': these do not form a Boolean algebra, but rather a
certain closed monoid). So here, we keep the general notion of a ``network'',
and recognize that the problem to be solved is to discover the relations,
and to uniformly, fairly assign probabilities to each relation. The
algorithm for discovering the relations, and to assigning appropriate
probabilities and metrics, is discussed in the next section. But first,
we continue expanding the linguistic horizon slightly.

\subsection{Deep Comprehension}

While the learning of syntactic and semantic relations is the primary
focus of the discussion here, the search for semantic structure must
not end there; more is possible. In particular, natural language generation
has a vital need for lexical functions, so that appropriate word-choices
can be made when vocalizing ideas. In order to truly understand text,
one also needs, as a minimum, to discern referential structure, and
sophisticated understanding requires discerning topics and themes.
Discerning such structure seems to be a bare minimum for what it means
to truly ``comprehend'' language. The aspects discussed here are
taken from Meaning-Text Theory (MTT), briefly summarized in the appendix.
Because these more abstract structures can again be viewed as graphical
relations, and as transformations on the structure of graphs, then
these too seem to be amenable to automated discovery.

A list of linguistic aspects that seem approachable are listed below.
We believe automated, unsupervised learning of these aspects is attainable,
building on top of the 'simpler' language language structures described
above. We are not aware of any prior work aimed at automatically learning
these, aside from relatively simple, unsophisticated (bag-of-words
style) efforts at topic categorization. Learning these may prove to
be extremely challenging, as the layered, recursive approach requires
that the earlier syntactic and semantic levels be relatively ``noise-free''
in order for more complex, more abstract structures to be discerned.
It is not yet clear that the earlier stages can achieve enough accuracy
to allow these later stages to proceed. What's more, two of these
aspects is where linguistics meets the reality of the external world,
and are arguably where ``understanding'' takes over from ``semantics''.
This is discussed further below.

So:
\begin{itemize}
\item Lexical functions should be learned.

\begin{itemize}
\item Lexical functions are (named) classes of predicate-argument relationships.
Thus, for example, the lexical function\noun{ Magn()} specifies a
list of appropriate words for expressing magnitude. One then has \noun{Magn}(\emph{rain})
== \emph{torrential}|\emph{hard}, \noun{Magn}(\emph{wind}) == \emph{strong}
and \noun{Magn}(\emph{emotion}) == \emph{hot.} Similarly, the subject
lexical function \noun{S\textsubscript{1}() }indicates authorship,
so \noun{S\textsubscript{1}(}\emph{crime}\noun{)} == \emph{perpetrator},\noun{
S\textsubscript{1}(}\emph{book}\noun{) }== \emph{author}. Lists of
synonyms and antonyms are also examples of lexical functions. In theories
of semantics, such as Meaning-Text Theory, dozens of lexical functions
are known and well-defined\cite{Mel'cuk1987,Kahane2003,Milicevic2006};
there may be more, but parsimony suggests that there cannot be thousands. 
\end{itemize}
\item Referential structure should be learned.

\begin{itemize}
\item References may be pronouns, or references to external objects. For
example, in ``\emph{Patricia went to the store. She bought a dress.}'',
the word ``\emph{she}'' refers to Patricia. In fact, the words ``\emph{dress}''
may also refer to a specific object in the observer's environment;
however, such a reference cannot be obtained by purely corpus-linguistic
methods. That said, in a prolonged discussion about a dress, it should
be possible to infer, without external cues, that the entire discussion
is about the same, singular dress, as opposed to a different dress
every time the word appears.
\item Referential structure requires a model of the external world, a model
of ``other'' and possibly a model of ``self''. That is, in order
to understand that the word ``\emph{dress}'' always refers to the
same object requires that there be a model of the world in which there
is one distinctive dress which can be the object of discussion. Likewise
many pronomial constructions require models of other actors, and also
a model of self as an actor, in order to be properly understood. 
\end{itemize}
\item Topic themes/communicative intent should be learned.

\begin{itemize}
\item Semantic-communicative structure captures the communicative intent;
it partitions a semantic structure graph into two parts: the 'theme'
(what is being talked about) and the 'rheme' (what is being said
about the theme). A semantic structure is just a graph of the semantic
relationships extracted from a sentence. Consider the (rather sophisticated)
sentence: ``The senator harshly criticized the Government for its
decision to increase income taxes''. This graph will contain semantic
relations such as \emph{criticize(X,Y,Z)}, which is to be understood
as '\emph{X criticizes Y for Z}'; this relation has three arrows from
'\emph{criticize}' to \emph{X}, \emph{Y} and \emph{Z}. Other arrows
link these in turn, to form a directed graph. One partitioning of
this graph is to take ``\emph{Government's decision to increase income
taxes}'' as the theme, and ``\emph{the senator's harsh criticism}''
as the rheme. This partitioning is appropriate when entire paragraphs
are devoted to the Government's decision; the Senator's criticism
is but one statement that pops up. 
\item Note that the above partitioning presumes that a rather sophisticated
referential structure has already been extracted: the theme should
appear in multiple sentences, and even in multiple paragraphs, and
should already have been identified as 'one and the same thing' across
these various appearances. 
\item This partitioning is not unique. One can also take ``the Senator''
as the theme, and ``criticism of government policy'' as the rheme.
Such a partitioning would be appropriate when reading the Senator's
biography. In other words, topics and themes cannot be discerned from
single sentences alone, but only become apparent from relationships
across many paragraphs.
\end{itemize}
\end{itemize}
The point here is that there are deeper structures in text, and that
these seem like they might be discernible in a mechanistic fashion.
This belief is built on the observation that these structures also
take the form of graphical relations. Whether the lower syntactic
and semantic relations are clean and coherent enough for these more
abstract structures to be discerned is unclear. 

At any rate, the referential structure is where language meets reality.
Natural language understanding, at this point, requires that a model
of the external world be accessible, so that referents can be attached
to the objects to which they might refer. This is again an act of
``parsing'', of joining the unattached ends of referential connectors
to the parts of the external world to which they might plausibly refer
to, and then checking the entire structure, by means of transitive
reasoning, for consistency. We use the word ``transitive'' here
in the same sense that one defines the ``transitive closure'' of
a relation: so, for example, if John says he was bitten by an animal,
and our model of the external world indicates that a dog was present,
one may then conclude that the ``animal'' refers to that particular
dog, by the application of a single joining relation: a dog is an
animal. At any rate, these observations should make clear that this
is where learning stops being about language, and instead turns into
something else. The something else is different and harder: it requires
models of the external world, and it requires reasoning, so that the
model of the external world can be manipulated to align with the topic
of conversation.

This also implies that the order of learning proposed here is reversed
from the normal direction in humans: children first construct models
of the external world based on visual inputs; these models are then
adorned with associated sounds, which eventually resolve into words,
due to their regularity. The regularity of syntax is the last to be
discerned, not the first. Doing the same in a disembodied context
is far, far harder: how does one discern that the external world consists
of persistent objects that have names (nouns) and are in changing
relationships to one-another (verbs)? Because of this reversed order,
it is quite possible that the learning proposed here will founder
shortly after the syntactic stage, falling far short of creating a
sensory perception model. At any rate, the proposal here entirely
omits any mechanism for constructing a model of the external world,
and correlating language with it.

\section{A Methodology for Unsupervised Language Learning from a Large Corpus}

The language learning approach presented here is novel in its overall
nature. Each part of it, however, draws on prior experimental and
theoretical research by others on particular aspects of language learning,
as well as on our own previous work building computational linguistic
systems. The goal is to assemble a system out of parts that are already
known to work well in isolation.

Prior published research, from a multitude of authors over the last
few decades, has already demonstrated how many of the items listed
above can be learnt in an unsupervised setting (see e.g. \cite{Yuret1998,Klein2004,Lin2001,Cohen2010,Poon2009,Mihalcea2007,Kart2013}
for relevant background). All of the previously demonstrated results,
however, were obtained in isolation, via research that assumed the
pre-existence of surrounding infrastructure far beyond what we assume
above. The approach proposed here may be understood as a combination,
generalization and refinement these techniques, to create a system
that can learn, more or less \emph{ab initio} from a large corpus,
with a final result of a working, usable natural language comprehension
system.

Thus, in some sense, the approach advocated here can be considered
to be a mash-up of techniques. However, a concomitant task is to formalize
the underlying mathematics of the undertaking, so that it becomes
clear what approximations are being taken, and what avenues remain
unexplored. Some fairly specific directions in this regard suggest
themselves, not the least of which is the need to write down appropriate
formulations for the distribution of probabilities, and the inequalities
that must hold in order for a learning event to occur.

Much of the prior research alluded to above makes use of probabilistic
arguments, usually with the implicit desire of treating unknowns fairly
and evenly. In Bayesian arguments, this amounts to a statement about
priors and assumptions about independence. In maximum-entropy methods,
the goal is to explicitly distribute unknown probabilities as evenly
as possible. Quite often, an approach is \emph{ad hoc}, with some
arbitrary but plausible metric providing a utility function that can
be maximized or minimized. Each approach has its pros and cons: independence
assumptions get Bayesian methods into trouble; the complexity of the
entropic partition function can sometimes make maximum entropy methods
intractable, while the \emph{ad hoc} nature of \emph{ad hoc} approaches
stymie a broader theoretical vision for the correct generalization
of a phenomenon.

The approach advocated below takes a pragmatic stance. That is, while
maximum entropy principles would seem to provide the correct theoretical
framework, they also require that it be clearly understood what is
being counted (so that the entropy can be correctly measured). Thus,
there is room for applying more traditional probabilistic reasoning,
and even ad hoc simplifications and short-cuts. As in most scientific
disciplines, progress here is best achieved by coupling experimental
exploration to theoretical and mathematical development.

\subsection{A High Level Perspective on Language Learning}

On an abstract conceptual level, the approach proposed here depicts
language learning as an instance of a general learning loop such as:
\begin{enumerate}
\item Observe structural relationships between linguistic entities (such
as words, or other entities described in previous sections). Find
frequently occurring and novel relationships: this can be done by
means of mutual information (for example), which adjusts for the novelty
of a (conditional) relationship by properly weighting it by the occurrence
frequency of the conditions in other contexts. That is, mutual information
is one good way to pluck novel relationships out of a sea of white
noise.
\item Give each structural relationship a (unique) name. The name is required
so that it can be counted and held and worked with. 
\item Treat each structure relationship as a constraint: relations that
are not seen are assumed to be prohibited. That is, there is no corpus
of grammatically incorrect sentences; rather, incorrectness is inferred
from silence. New inputs are grammatically ``parsable'' insofar
as they are consistent with the relationships (constraints) that have
been observed before.
\item Group together similar structural relationships. This is the application
of the law of parsimony, or of ``Occam's razor'': simply making
a list of all possible observed relationships, as suggested by steps
1 and 2, can result in very long lists; a shorter description is desired.
That is, one valid way of specifying a language is to provide an infinite
list of grammatically correct sentences. Such a list is not at all
compact; one wishes to group together similar sentences. Steps 1 and
2 suggest how to find the patterns about which one should group; there
remains the actual task of grouping, which is this step. Grouping
can be done using \emph{ad hoc} distance metrics, to discover similar
things. Grouping can also be done by entropy minimization (not maximization!)
methods: cutting a list down to \emph{N} items from 2\emph{N} items
by grouping reduces the entropy by $\log2=\log2N-\log N$.
\item For each such grouping make a category label, and add it to the lexis
of expected relations.
\item Return to Step 1, and restart observations, but this time, doing so
in terms of the known, expected relations. So, for example, if the
previous round of observations lead to the discovery of groupings
such as nouns and determiners, and the fact that these occur in immediate
proximity to one-another, then this should be taken as a ``known''
aspect of language. Armed with this relationship, perhaps other relationships
can now become clear, such as that nouns can sometimes be subjects,
and sometimes objects of verbs. With each iteration, new relationships
presumably emerge; however, they cannot become visible or clear until
all ``known'' aspects are already accounted for. Learning a set
of constraints sets a new baseline; deviations from the new baseline,
if they are strong enough, are candidates for new relations. Thus
the iterative nature of the algorithm.
\end{enumerate}
\noindent It stands to reason that the result of this sort of learning
loop, if successful, will be a hierarchically composed collection
linguistic relationships possessing the following

\parbox[c]{15cm}{%
 \textbf{Linguistic Coherence Property:} Linguistic entities are reasonably
well characterizable in terms of the compactly describable patterns
observable in their relationship with with other linguistic entities.%
}

This sort of property has observed to hold for many linguistic entities,
an observation dating back at least to Saussure \cite{Saussure1916}
and the start of structuralist linguistics. It is basically a fancier
way of saying that the meanings of words and other linguistic constructs,
may be found via their \textit{relationships} to other words and linguistic
constructs. We are not committed to structuralism as a theoretical
paradigm, and we have considerable respect for the aid that non-linguistic
information --such as the sensorimotor data that comes from embodiment
-- can add to language, as stressed in prior publications\cite{Goertzel2008a}.
However, the potential utility of non-linguistic information for language
learning does not imply the impossibility or infeasibility of learning
language from corpus data alone. It is inarguable that non-linguistic
relationships comprise a significant portion of the everyday meaning
of linguistic entities; but yet, redundancy is prevalent in natural
systems, and we believe that purely linguistic relationships may well
provide sufficient data for learning of natural languages. If there
are some aspects of natural language that cannot be learned via corpus
analysis, it seems difficult to identify what these aspects are via
armchair theorizing, and likely that they will only be accurately
identified via pushing corpus linguistics as far as it can go.

This generic learning process is a special case of the general process
of symbolization, described in \cite{Goertzel1994} and elsewhere
as a key aspect of general intelligence. In this process, a system
finds patterns in itself and its environment, and then symbolizes
these patterns via simple tokens or symbols that become part of the
system's native knowledge representation scheme (and hence parts of
its \textquotedbl{}metalanguage\textquotedbl{} for describing things
to itself). Having represented a complex pattern as a simple symbolic
token, it can then easily look at other patterns involving this patterns
as a component.

Note that in its generic format as stated above, the \textquotedbl{}language
learning loop\textquotedbl{} is not restricted to corpus based analysis,
but may also include extralinguistic aspects of usage patterns, such
as gestures, tones of voice, and the physical and social context of
linguistic communication. Linguistic and extra-linguistic factors
may come together to comprise \textquotedbl{}usage patterns.\textquotedbl{}
However, the restriction to corpus data does not necessarily denude
the language learning loop of its power; it merely restricts one to
particular classes of usage patterns, whose informativeness must be
empirically determined.

In principle, one might be able to create a functional language learning
system based only on a very generic implementation of the above learning
loops. In practice, however, biases toward particular sorts of usage
patterns can be very valuable in guiding language learning. In a computational
language learning context, it may be worthwhile to break down the
language learning process into multiple instances of the basic language
learning loops, each focused on different sorts of usage patterns,
and coupled with each other in specific ways. This is in fact what
we will propose here.

Specifically, the language learning process proposed here involves:
\begin{itemize}
\item One language learning loop for learning purely syntactic linguistic
relationships (such as link types and lexical entries, described above),
which are then used to provide input to a syntax parser.
\item One language learning loop for learning higher-level \textquotedbl{}syntactico-semantic\textquotedbl{}
linguistic relationships (such as semantic relationships, idioms,
and lexical functions, described above), which are extracted from
the output of the syntax parser.
\item One language learning loop for associating the resulting syntactico-semantic
relationships to a model of the external world.
\end{itemize}
\noindent These three loops are not independent of one-another; the
second loop can provide feedback to the first, regarding the correctness
of the extracted structures; then as the first loop produces more
correct, confident results, the second loop can in turn become more
confident in it's output. Likewise, the third loop selects proper
interpretations for the output of the second loop. In this sense,
the three loops attack the same sort of slow-convergence issues that
'deep learning' tackles in neural-net training.

The syntax parser itself, in this context, is used to extract \textit{directed
acyclic graphs} (dags), usually trees, from the graph of syntactic
relationships associated with a sentence. These dags represent parses
of the sentence. So the overall scope of the learning process proposed
here is to learn \textbf{a system of relationships that displays appropriate
coherence and that, when applied by an appropriate parser to the inputs
from the previous layer, will yield parse trees that reflect the information
content in the input.} The sensory input at the lowest layer is raw
text; the output is parse trees, which are then fed as input to the
next layer. The process is repeated, until a sufficiently abstract
form is obtained, such that it can be correlated with model of the
external world.

\subsection{Learning Syntax}

The process of learning syntax from a corpus may be understood fairly
directly in terms of entropy maximization. As a simple example, consider
the measurement of the entropy of the arrangement of words in a sentence.
To a fair degree, this can be approximated by the sum of the mutual
entropy between pairs of words. Yuret showed that by searching for
and maximizing this sum of entropies, one obtains a tree structure
that closely resembles that of a dependency parse\cite{Yuret1998}.
That is, the word pairs with the highest mutual entropy are more or
less the same as the arrows in a dependency parse, such as that shown
in figure \ref{fig:Dependency-and-Phrase-Structure}. Thus, an initial
task is to create a catalog of word-pairs with a large mutual entropy
(mutual information, or MI) between them. This catalog can then be
used to approximate the most-likely dependency parse of a sentence.

The link-types of such an unlabeled parse tree should be taken as
unique to each word-pair; that is, there is a unique link type to
connect any two (connectable) words. For a typical modern language,
this implies many millions of distinct link types; from the syntactic
viewpoint, this is intolerable, its an overly complex description
of language. The immediately obvious course of action is to somehow
group together different link types, reducing their number. But if
different words share a common link type, then perhaps the words should
also be grouped together into common classes. The result of such grouping
presumably results in the automated discovery of something similar
to part-of-speech groupings. So, for example, the computation of word-pair
MI is likely to reveal the following high-MI word pairs: ``\emph{big
car}'', ``\emph{fast car}'', \emph{``expensive car}'', \emph{``red
car}''. (Such word pairs have been previously observed in earlier
MI experiments.) It is intuitively obvious that one may group together
the words \emph{big}, \emph{expensive}, \emph{fast} and \emph{red}
into a single category, interpreted as modifiers to \emph{car}. But
how might the correctness of such a grouping be automatically verified?
The answer is relatively straight-forward: the same modifier grouping
can be observed acting on other nouns: \emph{e.g.} ``\emph{big bicycle}'',
``\emph{fast bicycle}'', \emph{etc.} Two effects are at play here:
a reinforcement of the correctness of the original grouping of modifiers,
but also the suggestion that perhaps cars and bicycles should be grouped
together. Superficially, it appears that one can discover two classes
of words from this example: modifiers and nouns; crudely put, parts
of speech. 

More importantly, the discovery comes about through a reduction in
the total number of syntax rules at play: rather than having to ``remember''
(log, record) millions of unique word pairs, it is sufficient to remember
a smaller number of word classes, and a smaller number of link types
to connect them. Entropy is defined as the logarithm in the total
number of states; complexity may be defined as the entropy in a set
of rules; such clustering is a reduction of the total complexity of
the syntactic description. This is biologically plausible: the human
mind does not maintain millions of syntactic relations, but an apparently
much smaller set, possibly in the hundreds, or less.

There is an interesting mathematical foundation for understanding
the role of link types; it comes from categorial grammar\cite{Kart2013}.
The link between two word classes carries a type; the type of that
link is \emph{defined} by these two classes. In this example, a link
between a modifier and a noun would be a type denoted as M\textbackslash{}N
in categorial grammar, M denoting the class of modifiers, and N the
class of nouns. In Link Grammar, this type name is replaced by a shorter
link name, without a slash, but is the same thing. (So, for this example,
the existing Link Grammar dictionaries use the A link for the M\textbackslash{}N
type, with A meant to conjure up 'adjective' as a mnemonic.) The short
link name is a boon for readability, as categorial grammars usually
have very complex-looking link-type names: \emph{e.g.} (NP\textbackslash{}S)/NP
for the simplest transitive verbs. The point being made here is that
typing and type theory\cite{HoTT2013} provide a good foundation for
dealing with the difficulty of 'naming things' discovered through
classification and clustering. The contents of a cluster is, in a
certain sense, \emph{ad hoc}, and based on what the texts that have
been ingested. The relations between clusters, however, are not: not
only are they dictated by language, but also come with an algebra
describing how they combine: this is type theory. In that sense, typing
seems to be an inherent part of language; type theory appears to provide
the correct formalization for discussing it. 

The Link Grammar dictionaries contain lists of disjuncts, not lists
of word-pairs. The last step of learning a workable grammar is then
to discover the disjuncts. This may be done by performing a minimum-spanning-tree
(MST) parse of input text\cite{McDonald2005,McDonald2006}, driven
entirely by the mutual information obtained between word-pairs. Given
that each link is implicitly labeled by the two words it joins, the
word connectors are trivially extracted as the link type, together
with a direction indicator. A disjunct is then simply the ordered
list of all of the connector that land on a given word. Thus, disjuncts
can be extracted on a sentence-by-sentence basis after a pass through
an MST parser. This then sets the stage for the next step of pattern
recognition.

Given a single word, appearing in many different sentences, one should
presumably find that this word only makes use of a relatively small,
limited set of disjuncts. It is then a counting exercise to determine
which disjuncts occur the most often for this word, and more, what
the disjuncts mutual information should be. The set of these disjuncts
then form this word's lexical entry. Similar to the discovery of high-MI
word pairs, this devolves into another \textquotedbl{}counting exercise\textquotedbl{}.
Because the structures being discovered are now subgraphs, instead
of word pairs, this is sometimes called ``frequent subgraph mining''.
This term is somewhat misleading: it is not the absolute frequency
of occurrence of the subgraph that is important, but its relative
(conditional) frequency, conditioned on the frequency of the other
parts of the graph that it connects to. The logarithm of the conditional
probability is called the relative entropy, or mutual information,
and so the counting exercise has again devolved into the computation
of MI of structures extracted from a corpus. An appendix gives formulas
to make these words precise; it is given because a proper definition
appears to be rare in the linguistic literature.

At this point, a second clustering step may be applied: its presumably
noticeable that many words use more-or-less the same disjuncts in
syntactic constructions. These can then be grouped into a common lexical
entry. Given that a different set of word groupings (into parts of
speech) was previously generated, one may ask: how does that grouping
compare to this grouping? Is it close, or can the groupings be refined?
If the groupings cannot be harmonized, then perhaps there is a certain
level of detail that was previously missed: perhaps one of the groups
should be split into several parts. Conversely, perhaps one of the
groupings was incomplete, and should be expanded to include more words.
Thus, there is a certain back-and-forth feedback between these different
learning steps, with later steps reinforcing or refining earlier steps,
forcing a new revision of the later steps. The precise refinement
of how this is to be done awaits experimental trials.

\subsubsection{Loose language}

A recognized difficulty with the direct application of Yuret's observation
(that the high-MI word-pair tree is essentially identical to the dependency
parse tree) is the flexibility of the preposition in the English language\cite{Klein2004}.
The preposition is so widely used, in such a large variety of situations
and contexts, that the mutual information between it, and any other
word or word-set, is rather low (is uniformly distributed, and thus
carries little information). The two-point, pair-wise mutual entropy
provides a poor approximation to what the English language is doing
in this particular case. It appears that the situation can be rescued
with the use of a three-point mutual information (a special case of
interaction information \cite{Bell2003}).

The discovery and use of such constructs is described in \cite{Poon2009}.
A similar, related issue can be termed ``the richness of the MV link
type in Link Grammar''. This one link type, describing verb modifiers
(which includes prepositions) can be applied in a very large class
of situations; as a result, discovering this link type, while at the
same time limiting its deployment to only grammatical sentences, may
prove to be a bit of a challenge. Even in the manually maintained
Link Grammar dictionaries, it can present a parsing challenge because
so many narrower cases can often be treated with an MV link. In summary,
some constructions in English are so flexible that it can be difficult
to discern a uniform set of rules for describing them; certainly,
pair-wise mutual information seems insufficient to elucidate these
cases.

Curiously, these more challenging situations occur primarily with
more complex sentence constructions. Perhaps the flexibility is associated
with the difficulty that humans have with composing complex sentences;
short sentences are almost 'set phrases', while longer sentences can
be a semi-grammatical jumble. In any case, some of the trouble might
be avoided by limiting the corpus to smaller, easier sentences at
first, perhaps by working with children's literature at first.

\subsubsection{Elaboration of the Syntactic Learning Loop}

We now reiterate the syntactic learning process described above in
a more systematic way. By getting more concrete, we also make certain
assumptions, and restrictions, some of which may end up getting changed
or lifted in the course of implementation and detailed exploration
of the overall approach proposed here. What is discussed in this section
is merely one simple, initial approach to concretizing the core language
learning loop we envision in a syntactic context.

Syntax, as we consider it here, involves the following basic entities:
\begin{itemize}
\item words 
\item categories of words 
\item \textquotedbl{}co-occurrence links\textquotedbl{}, each one defined
as (in the simplest case) an ordered pair or triple of words, labeled
with frequency counts and mutual information
\item \textquotedbl{}syntactic link types\textquotedbl{}, each one defined
by the two sets of words that are connected 
\item \textquotedbl{}disjuncts\textquotedbl{}, each one associated with
a particular word $w$, and consisting of an ordered set of link types
that connect to the word $w$. That is, each of these links contains
at least one word-pair containing $w$ as first or second argument.
(This nomenclature here comes from Link Grammar; each disjunct is
a conjunction of link types. A word is associated with a set of disjuncts.
In the course of parsing, one must choose between the multiple disjuncts
associated with a word, to fulfill the constraints required of an
appropriate parse structure.) 
\end{itemize}
An elementary version of the basic syntactic language learning loop
described above would take the form.
\begin{enumerate}
\item Search for high-MI word pairs. Define an initial set of word-pair
link types as the given co-occurrence links.
\item Cluster words into categories based on the similarity of their associated
usage links 

\begin{itemize}
\item Note that this will likely be a tricky instance of clustering, and
classical clustering algorithms may not perform well. One interesting,
less standard approach would be to use OpenCog's MOSES algorithm to
learn an array of program trees, each one serving as a recognizer
for a single cluster. 
\end{itemize}
\item Define initial syntactic link types from categories that are joined
by large bundles of usage links 

\begin{itemize}
\item That is, if the words in category $C_{1}$ have a lot of usage links
to the words in category $C_{2}$, then create a syntactic link type
whose elements are $(w_{1},w_{2})$, for all $w_{1}\in C_{1},w_{2}\in C_{2}$.
Remove the word-pair link types associated with $(w_{1},w_{2})$,
as these are now all subsumed by the new link type.
\end{itemize}
\item Associate each word with an extended set of usage links, consisting
of: its existing usage links, plus the syntactic links that one can
infer for it based on the categories the word belongs to. These are
the ``disjuncts'' of Link Grammar. Typically, determiners and adjectives
have just one link (linking to the modified noun), nouns have one
or two links (one to an adjective, one to a verb) while verbs typically
have three or four links (one to a subject, one to an object, possibly
a link to a particle or preposition or other adverb, and one to the
head). 

\begin{itemize}
\item For example, suppose $cat\in C_{1}$ and $C_{1}$ has syntactic link
$L_{1}$. Suppose $(cat,eat)$ and $(dog,run)$ are both in $L_{1}$.
Then if there is a sentence \textquotedbl{}The cat likes to run\textquotedbl{},
the link $L_{1}$ lets one infer the syntactic link $\mathrm{cat}\overset{L_{1}}{\rightarrow}\mathrm{run}$.
The corpus is re-scanned to obtain the frequency of this syntactic
link, as well as its mutual information (logarithm of the conditional
probability).
\item Given the sentence \textquotedbl{}The cat likes to run in the park,\textquotedbl{}
a chain of syntactic links such as $\mathrm{cat}\overset{L_{1}}{\rightarrow}\mathrm{run}\overset{L_{2}}{\rightarrow}\mathrm{park}$
may be constructed. 
\end{itemize}
\item Look for commonality between disjuncts. This may indicate clusterings
of words or link-types that were previously missed; alternately, these
may indicate that previous clusterings were excessively aggressive. 
\item Return to Step 2, but using the extended set of usage links produced
in Step 4, with the goal of refining clusters, the set of link types,
and the set of disjuncts for accuracy. Initially, all categories contain
one word each, and there is a unique link type for each pair of categories.
This is an inefficient representation of language, and so the goal
of clustering is to have a relatively small set of clusters and link
types, with many words/word-pairs assigned to each. This can be done
by maximizing the sum of the logarithms of the sizes of the clusters
and link types; that is, by maximizing entropy. Since the category
assignments depend on the link types, and \emph{vice versa}, a (very?)
large number of iterations of the loop are likely to be required.
Based on the current Link Grammar English dictionaries, one expects
to discover hundreds of link types (or more, depending on how subtypes
are counted), and perhaps a thousand word clusters (most of these
corresponding to irregular verbs and idiomatic phrases).
\end{enumerate}
Many variants of this same sort of process are conceivable, and it's
currently unclear what sort of variant will work best. But this \textit{kind}
of process is what one obtains when one implements the basic language
learning loop described above on a purely syntactic level.

How might one integrate semantic understanding into this syntactic
learning loop? Once one has semantic relationships associated with
a word, one uses them to generate new \textquotedbl{}usage links\textquotedbl{}
for the word, and includes these usage links in the algorithm from
Step 1 onwards. This may be done in a variety of different ways, and
one may give different weightings to syntactic versus semantic usage
links, resulting in the learning of different links.

The above process would produce a large set of syntactic links between
words. We then find a further series of steps. These may be carried
out concurrently with the above steps, as soon as Step 4 has been
reached for the first time.
\begin{enumerate}
\item Given the set of disjuncts, one carries out parsing using a process
such as link parsing or word grammar parsing, thus arriving at a set
of parses for the sentences in one's reference corpus. Alternative
parses may be ranked according to the total mutual information, summed
over all disjuncts. Each parse may be viewed as a directed acyclic
graph (dag), usually a tree, with words at the nodes and syntactic-link
type labels on the links. 
\item This allows a different set of statistics to be gathered for each
disjunct: how often it proves \emph{actually useful} during (link-typed)
parsing. That is, the initial probabilities and entropies for the
disjuncts essentially followed from how often they are employed by
an MST parser, which generates a spanning tree essentially without
regard to the link types. Re-parsing, this time with actual link type
agreement enforced, will presumably give similar parses, but presumably
more accurate ones. In addition, the usage probabilities will change
as a result. \\
\\
In particular, forcing link type agreement might cause some words
to be missed in the parse. For an MST parse, this is not an issue:
one simply hunts for some high-MI connection, and attaches to that.
With link types, there may be no valid linkage at all. This suggests
the existence of a problem with the current link type and disjunct
assignments: these are somehow incomplete, if they fail to link all
the words in the sentence. In essence, link parsing is stricter than
MST parsing; the strictness is a source of feedback for validating
the grammar.
\item One can now return to Step 2 using the new probabilities, which should
suggest new and refined clusters. 
\end{enumerate}
Several subtleties have been ignored in the above, such as the proper
discovery and treatment of idiomatic phrases, the discovery of sentence
boundaries, the handling of embedded data (price quotes, lists, chapter
titles, \emph{etc.}) as well as the potential speed bump that are
prepositions. Fleshing out the details of this loop into a workable,
efficient design is the primary engineering challenge. This will take
significant time and effort.

\subsection{Learning Semantics}

Syntactic relationships provide only the shallowest interpretation
of language; semantics comes next. One may view semantic relationships
(including semantic relationships close to the syntax level, which
we may call \textquotedbl{}syntactico-semantic\textquotedbl{} relationships)
as ensuing from syntactic relationships, via a similar but separate
learning process to the one proposed above. Just as our approach to
syntax learning is heavily influenced by our work with Link Grammar,
our approach to semantics is heavily influenced by our work on the
RelEx system \cite{RelEx,Lian2010,Goertzel2006,Lian2012}, which maps
the output of the Link Grammar parser into a more abstract, semantic
form. Prototype systems \cite{Goertzel2010,Lian2012} have also been
written mapping the output of RelEx into even more abstract semantic
form, consistent with the semantics of the Probabilistic Logic Networks
\cite{PLN} formalism as implemented in the OpenCog \cite{Goertzel2008}
framework. These systems are largely based on hand-coded rules, and
thus not in the spirit of language learning pursued in this proposal.
However, they display the same \textit{structure} that we assume here;
the difference being that here we specify a mechanism for learning
the linguistic content that fills in the structure via unsupervised
corpus learning, obviating the need for hand-coding.

Specifically, we suggest that discovery of semantic relations can
proceed in a manner similar to the unsupervised discovery of synonyms,
such as that described in \cite{Lin2001}, or it's generalization
from 2-point relations to 3-point and N-point relations, as described
in \cite{Poon2009}. These mechanisms allow the automatic, unsupervised
recognition of synonymous phrases, such as ``Texas borders on Mexico''
and ``Mexico is Texas neighbor'', to extract the general semantic
relation \emph{next\_to(X,Y)}, and the fact that this relation can
be expressed in one of several different ways.

Simplistically stated, the idea is that semantic learning can proceed
by scanning the corpus for sentences that use similar or the same
words, yet employ them in a different order, or have point substitutions
of single words, or of small phrases. Sentences which are very similar,
or identical, save for one word, offer up candidates for synonyms,
or sometimes antonyms. Sentences which use the same words, but in
seemingly different syntactic constructions, are candidates for synonymous
sentences. These may be used to extract semantic relations: the recognition
of sets of different syntactic constructions that carry the same meaning.
In practice, the comparisons and search for similarity is not made
on the raw text strings, but on the parsed forms of the sentences,
so as to avoid issues of word alignment during comparison. Parsing
establishes a graph that provides a context for differences in subgraphs. 

In essence, similar parse structures must be recognized, and then
word and parse-tree differences between other-wise similar parse graphs
are compared. There are two primary challenges: how to recognize similar
graphs, and how to assign probabilities.

The work of \cite{Poon2009} articulates solutions to both challenges.
For the first, it describes a general framework in which relations
such as \emph{next\_to(X,Y) }can be understood as lambda-expressions
$\lambda x\lambda y.\mbox{next\_to}(x,y)$, so that one can employ
first-order logic constructions in place of graphical representations.
This is partly a notational trick; it just shows how to split up input
syntactic constructions into atoms and terms, forming a term algebra
with signature\cite{Baader1999,Hodges1997}. For the second challenge,
they show how probabilities can be assigned to the atoms and terms,
by making explicit use of the notions of conditional random fields
(or rather, a certain special case, termed Markov Logic Networks).
Conditional random fields, or Markov networks, are a mathematical
formalism that describes how entropy can be uniformly distributed
across a graphical network where edges and verticies may both be typed,
and range over a domain of values. As such, this generalizes a basic,
fundamental theorem from information theory, that the probability
distribution that most evenly distributes unknowns or priors is the
same as the probability distribution that maximizes the entropy\cite{Ash1965}.
Unfortunately, the general theory has several drawbacks: it is quite
abstract and dense, and algorithmically, it falls into the NP-hard
class of problems. 

The procedures described in \cite{Lin2001} thus provide a much simpler,
easier-to-understand introduction to how semantic information can
be extracted. With that simplicity comes two faults: a lack of proper
mathematical grounding means that it is not clear how to generalize
the work to sub-graphs of arbitrary shape (which is provided by Poon
\& Domingos), nor is the generalized probabilistic framework articulated.
Instead, Lin gets by with a number of \emph{ad hoc} metrics used to
measure semantic similarity. This may be a reasonable approach: the
\emph{ad hoc} similarity metrics have the side effect of taking the
NP-hard maximum entropy algorithm and replacing it with a simpler,
more rapidly convergent method. 

The above can be used to extract synonymous constructions, and, in
this way, semantic relations. However, neither of the above references
deal with distinguishing different meanings for a given word. That
is, while \emph{eats(X,Y)} might be a learnable semantic relation,
the sentence ``\emph{He ate it}'' does not necessarily justify its
use. Of course: ``\emph{He ate it}'' is an idiomatic expression
meaning ``\emph{he crashed}'', which should be associated with the
semantic relation \emph{crash}(X), not \emph{eat(X,Y)}. There are
global textual clues that this may be the case: trouble resolving
the reference ``\emph{it}'', and a lack of mention of foodstuffs
in neighboring sentences. A viable yet simple algorithm for the disambiguation
of meaning is offered by the Mihalcea algorithm\cite{Mihalcea2004,Mihalcea2005,Mihalcea2007}.
This is an application of the (Google) PageRank algorithm to word
senses, taken across words appearing in multiple sentences. The premise
is that the correct word-sense is the one that is most strongly supported
by senses of nearby words; a graph between word senses is drawn, and
then solved as a Markov chain. In the original formulation, word senses
are defined by appealing to WordNet, and affinity between word-senses
is obtained via one of several similarity measures. Neither of these
can be applied in learning a language \emph{de novo} (one has neither
a WordNet for the language, nor any similarity measures). Instead,
these must both be deduced by clustering and splitting, again. So,
for example, it is known that word senses correlate fairly strongly
with disjuncts (based on authors unpublished experiments),%
\footnote{This can be understood in a simple, intuitive fashion. Traditional
dictionary entries are grouped according to the part of speech of
a word: different parts of speech are associated with different word
senses. The Link Grammar disjunct is like an extremely fine-grained
part of speech: it distinguishes not only between noun and verb, but
also between the contexts in which it is used (transitive, ditransitive,
with or without modifiers, quantifiers, determiners, particles, \emph{etc}.)
That word senses might correlate with this fine-grained part of speech
should come as no surprise. Such correlation is not unique to Link
Grammar; it should be directly observable in any dependency grammar.
The correlation might be harder to detect in phrase-structure grammars,
since lexical entries are not words, but phrase structures, and thus
its not obvious how to correlate word senses to phrase structures.%
} and thus, a reasonable first cut is to presume that every different
disjunct in a lexical entry conveys a different meaning, until proved
otherwise. The above-described discovery of synonymous phrases can
then be used to group different disjuncts into a single ``word sense''.
Disjuncts that remain ungrouped after this process are already considered
to have distinct senses, and so can be used as distinct senses in
the Mihalcea network.

Sense similarity measures can then be developed by using the above-discovered
senses, and measuring how well they correlate across different texts.
That is, if the word ``\emph{bell}'' occurs multiple times in a
sequence of paragraphs, it is reasonable to assume that each of these
occurrences are associated with the same meaning. Thus, each distinct
disjunct for the word ``\emph{bell}'' can then be presumed to still
convey the same sense. One now asks, what words co-occur with the
word ``\emph{bell}''? The frequent appearance of ``\emph{chime}''
and ``\emph{ring}'' can and should be noted. In essence, one is
once-again computing word-pair mutual information, except that now,
instead of limiting word-pairs to be words that are near each other,
they can instead involve far-away words, several sentences apart.
One can then expand the word sense of ``\emph{bell}'' to include
a list of co-occurring words (and indeed, this is the slippery slope
leading to set phrases and eventually idioms).

Failures of co-occurrences can also further strengthen distinct meanings.
Consider ``\emph{he chimed in}'' and ``\emph{the bell chimed}''.
In both cases, \emph{chime} is a verb. In the first sentence, \emph{chime}
carries the disjunct \texttt{S- \& K+} (here, \texttt{K+} is the standard
Link Grammar connector to particles) while the second has only the
simpler disjunct \texttt{S-}. Thus, based on disjunct usage alone,
one already suspects that these two have a different meaning. This
is strengthened by the lack of occurrence of words such as ``\emph{bell}''
or ``\emph{ring}'' in the first case, with a frequent observation
of words pertaining to talking.

There is one final trick that must be applied in order to get reasonably
rapid learning; this can be loosely thought of as ``the sigmoid function
trick of neural networks'', though it may also be manifested in other
ways not utilizing specific neural net mathematics. The key point
is that semantics intrinsically involves a variety of uncertain, probabilistic
and fuzzy relationships; but in order to learn a robust hierarchy
of semantic structures, one needs to iteratively crispen these fuzzy
relationships into strict ones.

In much of the above, there is a recurring need to categorize, classify
and discover similarity. The naivest means of doing so is by counting,
and applying basic probability (Bayesian, Markovian) to the resulting
counts to deduce likelihoods. Unfortunately, such formulas distribute
probabilities in essentially linear ways (\emph{i.e.} form a linear
algebra), and thus have a rather poor ability to discriminate or distinguish
(in the sense of receiver operating characteristics, of discriminating
signal from noise). Consider the last example: the list of words co-occurring
with \emph{chime}, over the space of a few paragraphs, is likely to
be tremendous. Most of this is surely noise. There is a trick to over-coming
this that is deeply embedded in the theory of neural networks, and
yet completely ignored in probabilistic (Bayesian, Markovian) networks:
the sigmoid function. The sigmoid function serves to focus attention
on a single stimulus, and elevate its importance, and, at the same
time, strongly suppress all other stimuli. In essence, the sigmoid
function looks at two probabilities, say 0.55 and 0.45, and says ``let's
pretend the first one is 0.9 and the second one is 0.1, and move forward
from there''. It builds in a strong discrimination to all inputs.
In standard, text-book probability theory, such discrimination is
utterly unwarranted; it runs counter to probability theory. However,
applying strong discrimination to learning can help speed learning
by converting certain vague impressions into certainties. These certainties
may be correct or incorrect; it is the task of learning to distinguish
the two. The point here is that this non-linear behavior provides
a kind of amplification, which allows vague impressions to be converted
into certainties that can then be built upon to obtain additional
certainties.

Thus, in all of the above efforts to gauge the similarity between
different things, it is useful to have a sharp yes/no answer, rather
than a vague muddling with likelihoods. In some of the above-described
algorithms, this sharpness is already built in: so, Yuret approximates
the mutual information of an entire sentence as the sum of mutual
information between word pairs: the smaller, unlikely corrections
are discarded. Clearly, they must also be revived in order to handle
prepositions. Something similar must also be done in the extraction
of synonymous phrases, semantic relations, and meaning; the domain
is that much likelier to be noisy, and thus, the need to discriminate
signal from noise that much more important.

\subsubsection{Elaboration of the Semantic Learning Loop}

We now provide a more detailed elaboration of a simple version of
the general semantic learning process described above. The same caveat
applies here as in our elaborated description of syntactic learning
above: the specific algorithmic approach outlined here is a simple
instantiation of the general approach we have in mind, which may well
require refinement based on lessons learned during experimentation
and further theoretical analysis.

One way to do semantic learning, according to the approach outlined
above, is as follows:
\begin{enumerate}
\item An initial semantic corpus is posited, whose elements are parse graphs
produced by the syntactic process described earlier.
\item A semantic relationship set (or \textbf{rel-set}) is computed from
the semantic corpus, via calculating the frequent (or otherwise statistically
informative, i.e. high MI) subgraphs occurring in the elements of
the corpus. Each node of such a subgraph may contain a word, a category
or a variable; each node may be typed by the list of connections it
is allowed to make. The links of the subgraph are labeled with (syntactic,
or semantic) link types. Each parse graph is annotated with the semantic
graphs associated with the words it contains (explicitly: each word
in a parse graph may be linked via a ReferenceLink to each variable
or literal with a semantic graph that corresponds to that word in
the context of the sentence underlying the parse graph.) 

\begin{itemize}
\item For instance, the link combination $\mathrm{v_{1}}\overset{S}{\longrightarrow}\mathrm{v_{2}}\overset{O}{\longrightarrow}\mathrm{v_{3}}$
may commonly occur (representing the standard Subject-Verb-Object
(SVO) structure).
\item For this example, the sentence \emph{``the rock broke the window}''
would result in links of the form $\mathrm{rock}\xrightarrow{ReferenceLink}\mathrm{v_{1}}$
connecting the word-instance nodes (the \textquotedbl{}\emph{rock}\textquotedbl{}
node) in the parse structure with variable nodes (such as $v_{1}$)
in this associated semantic subgraph. 
\end{itemize}
\item Rel-sets are divided into categories based on the similarities of
their associated semantic graphs. 

\begin{itemize}
\item This division into categories manifests the sigmoid-function-style
crispening mentioned above. Each rel-set will have similarities to
other rel-sets, to varying fuzzy degrees. Defining specific categories
turns a fuzzy web of similarities into crisp categorial boundaries;
which involves some loss of information, but also creates a simpler
platform for further steps of learning. 
\item Two semantic graphs may be called \textquotedbl{}associated\textquotedbl{}
if they have a nonempty intersection. The intersection determines
the type of association involved. Similarity assessment between graphs
$\mathcal{G}_{1}$ and $\mathcal{G}_{2}$ may involve estimation of
which graphs $\mathcal{G}_{1}$ and $\mathcal{G}_{2}$ are associated
with in which ways. Intersection is done by finding the largest common
subgraph.
\item For instance, \textquotedbl{}The cat ate the dog\textquotedbl{} and
\textquotedbl{}The frog was eaten by the walrus\textquotedbl{} both
represent the semantic structure \textit{eat(cat,dog)} in two different
ways. In link parser terminology, they do so respectively via the
subgraphs $\mathcal{G}_{1}=\mathrm{v_{1}}\overset{S}{\longrightarrow}\mathrm{v_{2}}\overset{O}{\longrightarrow}\mathrm{v_{3}}$
and $\mathcal{G}_{2}=\mathrm{v_{1}}\overset{S}{\longrightarrow}\mathrm{v_{2}}\overset{P}{\longrightarrow}\mathrm{v_{3}}\overset{MV}{\longrightarrow}\mathrm{v_{4}}\overset{J}{\longrightarrow}\mathrm{v_{5}}$.
These two semantic graphs will have a lot of the same associations.
For instance, in our corpus we may have \textquotedbl{}The big cat
ate the dog in the morning\textquotedbl{} (including $\mathrm{big}\overset{A}{\longrightarrow}\mathrm{cat}$)
and also \textquotedbl{}The big frog was eaten by the walrus in the
morning\textquotedbl{} (including $\mathrm{big}\overset{A}{\longrightarrow}\mathrm{frog}$),
meaning that $\mathrm{big}\overset{A}{\longrightarrow}\mathrm{v_{5}}$
is a subgraph commonly associated with both $\mathcal{G}_{1}$ and
$\mathcal{G}_{2}$. Due to having many commonly associated graphs
like this, $\mathcal{G}_{1}$ and $\mathcal{G}_{2}$ are likely to
be associated to a common cluster. 
\item As in syntactic parsing, a reasonable metric for clustering can be
obtained by applying pressure to reduce the overall complexity of
the system. Overall complexity is obtained by counting: summing the
total number of rules, relations, classes and class sizes needed to
capture the content of the language. Insofar as the logarithm of a
count is the entropy, this is an exercise in entropy minimization.
The goal is to prune away relations which never occur (because they
are semantic nonsense: \emph{``colorless green ideas}''), leaving
behind only those which can be used to express legitimate facts.
\end{itemize}
\item Nodes referring to these categories are added to the parse graphs
in the semantic corpus. Most simply, a category node $C$ is assigned
a link of type $L$ pointing to another node $x$, if any element
of $C$ has a link of type $L$ pointing to $x$. (More sophisticated
methods of assigning links to category nodes may also be worth exploring.) 

\begin{itemize}
\item If $\mathcal{G}_{1}$ and $\mathcal{G}_{2}$ have been assigned to
a common category $C$, then \textquotedbl{}I believe the pig ate
the horse\textquotedbl{} and \textquotedbl{}I believe the law was
invalidated by the revolution\textquotedbl{} will both appear as instantiations
of the graph $\mathcal{G}_{3}=\mathrm{v_{1}}\overset{S}{\longrightarrow}\mathrm{believe}\overset{CV}{\longrightarrow}\mathrm{C}$.
This $\mathcal{G}_{3}$ is compact because of the recognition of $C$
as a cluster, leading to its representation as a single symbol. The
recognition of $\mathcal{G}_{3}$ will occur in Step 2 the next time
around the learning loop. 
\end{itemize}
\item Return to Step 2, with the newly enriched semantic corpus. 
\end{enumerate}
As noted earlier, these semantic relationships may be used in the
syntactic phase of language understanding in two ways:
\begin{itemize}
\item Semantic graphs associated with words may be considered as \textquotedbl{}usage
links\textquotedbl{} and thus included as part of the data used for
syntactic category formation.
\item During the parsing process, full or partial parses leading to higher-probability
semantic graphs may be favored.
\end{itemize}

\section{The Importance of Incremental Learning}

Note that the above sequence of learning steps vaguely resembles the
layering of ``deep learning'', or of hierarchical modeling. That
is, learning must occur at several levels at once, each reinforcing,
and making use of results from another. Link types cannot be identified
until word clusters are found, and word clusters cannot be found until
word-pair relationships are discovered. However, once link-types are
known, these can be then used to refine clusters and the selected
word-pair relations.

The learning process described here builds up complex syntactic and
semantic structures from simpler ones. To start it, all one needs
are basic before and after relationships derived from a corpus. Everything
else is built up from there, given the assumption of appropriate syntactic
and semantic formalisms and a semantics-guided syntax parser.

However, for this bootstrapping learning to work well, one will likely
need to begin with simple language, so that the semantic relationships
embodied in the text are not that far removed from the simple before/after
relationships. The complexity of the texts may then be ramped up gradually.
For instance, the needed effect might be achieved via sorting a very
large corpus in order of increasing reading level.

\section{Conclusion}

We propose a general algorithm through which syntax and semantics
can be induced from a large text corpus. The algorithm builds on established
ideas and demonstrated algorithms from the linguistic and machine
learning fields; the primary novelty proposed here is that these are
mature enough to be able to be hooked together, so that the results
from induction at a lower layer can be used to provide input to a
higher layer, and that, conversely, higher layers can be used to guide
learning in the lower layers.

The ordering of the layers proposed here is reversed from the ordering
one might expect in embodied cognition, where one might first seek
to associate single words with non-verbal sensory inputs, eventually
building a model of the external world labeled with nouns for persistent
objects, and verbs for actions. Such labeling provides the foundation
for semantics in embodied cognition; syntax is learned only later,
to encode semantics that single words alone cannot capture. Here,
we reverse the process, attempting to learn syntax first, and only
later the semantics, and possible a world-model. Whether this is feasible
is remains unclear: neither form of learning, in one direction, or
the other, has ever been demonstrated. There are certainly any number
of traps and pitfalls in the descriptions given above. Nonetheless,
we believe that the state of the art in both mathematical theory,
and in the power of computer systems is sufficiently advanced that
this can be attempted.

Initial experiments to test some of these hypothesis have been performed
by the authors over the last number of years. Work has begun to implement
the ideas proposed here. The work is slow, as it is under-staffed;
the project described here is large, requiring man-years to demonstrate
even a vague prototype, and possibly dozens or hundreds to develop
a polished system, with a fully developed theory supported by detailed
published experiments.

Current work can be found in the OpenCog codebase: \texttt{http://github.com/opencog/opencog/}
most specifically, in the \texttt{opencog/nlp/learn} subdirectory,
with supporting code in other directories. The overall OpenCog project
is primarily focused on embodied cognition, and so both directions
are being explored: learning language via embodied models of the external
world, as well as learning language from corpus analysis, as described
here.

\bibliographystyle{alpha}
\bibliography{lang}

\newcommand{\etalchar}[1]{$^{#1}$}
\begin{thebibliography}{MPRH05}

\bibitem[Ash65]{Ash1965}
Robert~B. Ash.
\newblock {\em Information Theory}.
\newblock Dover Publications, 1965.

\bibitem[Bel03]{Bell2003}
Anthony~J. Bell.
\newblock The co-information lattice.
\newblock {\em Somewhere or other}, 2003.

\bibitem[BN99]{Baader1999}
Franz Baader and Tobias Nipkow.
\newblock {\em Term rewriting and all that}.
\newblock Cambridge University Press, 1999.

\bibitem[CS10]{Cohen2010}
Shay~B. Cohen and Noah~A. Smith.
\newblock Covariance in unsupervised learning of probabilistic grammars.
\newblock {\em Journal of Machine Learning Research}, 11:3117--3151, 2010.

\bibitem[dS77]{Saussure1916}
Ferdinand de~Saussure.
\newblock {\em Course in General Linguistics}.
\newblock Fontana/Collins, 1977.
\newblock Orig. published 1916 as ''Cours de linguistique g\'en\'erale''.

\bibitem[Gib98]{Gibson1998}
Edward Gibson.
\newblock Linguistic complexity: locality of syntactic dependencies.
\newblock {\em Cognition}, 68:1--76, 1998.

\bibitem[GIGH08]{PLN}
B.~Goertzel, M.~Ikle, I.~Goertzel, and A.~Heljakka.
\newblock {\em Probabilistic Logic Networks}.
\newblock Springer, 2008.

\bibitem[Goe94]{Goertzel1994}
Ben Goertzel.
\newblock {\em Chaotic Logic}.
\newblock Plenum, 1994.

\bibitem[Goe08]{Goertzel2008a}
Ben Goertzel.
\newblock A pragmatic path toward endowing virtually-embodied ais with
  human-level linguistic capability.
\newblock IEEE World Congress on Computational Intelligence (WCCI), 2008.

\bibitem[GPA{\etalchar{+}}10]{Goertzel2010}
Ben Goertzel, Cassio Pennachin, Samir Araujo, Ruiting Lian, Fabricio Silva,
  Murilo Queiroz, Welter Silva, Mike Ross, Linas Vepstas, and Andre Senna.
\newblock A general intelligence oriented architecture for embodied natural
  language processing.
\newblock In {\em Proceedings of the Third Conference on Artificial General
  Intelligence}. Springer, 2010.

\bibitem[GPPG06]{Goertzel2006}
Ben Goertzel, Hugo Pinto, Cassio Pennachin, and Izabela~Freire Goertzel.
\newblock Using dependency parsing and probabilistic inference to extract
  relationships between genes, proteins and malignancies implicit among
  multiple biomedical research abstracts.
\newblock In {\em Proc. of Bio-NLP 2006}, 2006.

\bibitem[HG08]{Goertzel2008}
David Hart and Ben Goertzel.
\newblock Opencog: A software framework for integrative artificial general
  intelligence.
\newblock In {\em Proceedings of the First Conference on Artificial General
  Intelligence}. IOS Press, 2008.

\bibitem[Hod97]{Hodges1997}
Wilfred Hodges.
\newblock {\em A Shorter Model Theory}.
\newblock Cambridge University Press, 1997.

\bibitem[Hud84]{Hud84}
Richard Hudson.
\newblock {\em Word Grammar}.
\newblock Oxford: Blackwell, 1984.

\bibitem[Hud07]{Hud07}
Richard Hudson.
\newblock {\em Language Networks: The New Word Grammar}.
\newblock Oxford Linguistics, 2007.

\bibitem[iC06]{Ferrer2006}
R.~Ferrer i~Cancho.
\newblock Why do syntactic links not cross?
\newblock {\em EPL (Europhysics Letters)}, 76(6):1228--1234, 2006.

\bibitem[Kah03]{Kahane2003}
Sylvain Kahane.
\newblock The meaning-text theory.
\newblock {\em Dependency and Valency. An International Handbook of
  Contemporary Research}, 1:546--570, 2003.

\bibitem[KM04]{Klein2004}
Dan Klein and Christopher~D. Manning.
\newblock Corpus-based induction of syntactic structure: Models of dependency
  and constituency.
\newblock In {\em ACL '04 Proceedings of the 42nd Annual Meeting on Association
  for Computational Linguistics}, pages 479--486. Association for Computational
  Linguistics, 2004.

\bibitem[KSPC13]{Kart2013}
Dimitri Kartsaklis, Mehrnoosh Sadrzadeh, Stephen Pulman, and Bob Coecke.
\newblock Reasoning about meaning in natural language with compact closed
  categories and frobenius algebras.
\newblock 2013.

\bibitem[LGE10]{Lian2010}
Ruiting Lian, Ben Goertzel, and Al~Et.
\newblock Language generation via glocal similarity matching.
\newblock {\em Neurocomputing}, 2010.

\bibitem[LGK{\etalchar{+}}12]{Lian2012}
Ruiting Lian, Ben Goertzel, Shujing Ke, Jade OÕNeill, Keyvan Sadeghi, Simon
  Shiu, Dingjie Wang, Oliver Watkins, and Gino Yu.
\newblock Syntax-semantic mapping for general intelligence: Language
  comprehension as hypergraph homomorphism, language generation as constraint
  satisfaction.
\newblock In {\em Artificial General Intelligence: Lecture Notes in Computer
  Science Volume 7716}. Springer, 2012.

\bibitem[Liu08]{Liu2008}
Haitao Liu.
\newblock Dependency distance as a metric of language comprehension difficulty.
\newblock {\em Journal of Cognitive Science}, 9(2):159--191, 2008.

\bibitem[LP01]{Lin2001}
Dekang Lin and Patrick Pantel.
\newblock Dirt: Discovery of inference rules from text.
\newblock In {\em Proceedings of the Seventh ACM SIGKDD International
  Conference on Knowledge Discovery and Data Mining (KDD'01)}, pages 323--328.
  ACM Press, 2001.

\bibitem[Mih05]{Mihalcea2005}
Rada Mihalcea.
\newblock Unsupervised large-vocabulary word sense disambiguation with
  graph-based algorithms for sequence data labeling.
\newblock In {\em HLT '05: Proceedings of the conference on Human Language
  Technology and Empirical Methods in Natural Language Processing}, pages
  411--418, Morristown, NJ, USA, 2005. Association for Computational
  Linguistics.

\bibitem[Mil06]{Milicevic2006}
Jasmina Mili{\'c}evi{\'c}.
\newblock A short guide to the meaning-text linguistic theory.
\newblock {\em Journal of Koralex}, (8):187--233, 2006.

\bibitem[MLP06]{McDonald2006}
Ryan McDonald, Kevin Lerman, and Fernando Pereira.
\newblock Multilingual dependency analysis with a two-stage discriminative
  parser.
\newblock In {\em CoNLL-X '06: Proceedings of the Tenth Conference on
  Computational Natural Language Learning}, pages 216--220, Morristown, NJ,
  USA, 2006. Association for Computational Linguistics.

\bibitem[MP87]{Mel'cuk1987}
Igor~A. Mel'{\v{c}}uk and Alain Polguere.
\newblock A formal lexicon in meaning-text theory.
\newblock {\em Computational Linguistics}, 13:261--275, 1987.

\bibitem[MPRH05]{McDonald2005}
Ryan McDonald, Fernando Pereira, Kiril Ribarov, and Jan Haji{\v{c}}.
\newblock Non-projective dependency parsing using spanning tree algorthms.
\newblock In {\em HLT-EMNLP 05 Proceedings of the conference on Human Language
  Technology and Empirical Methods in Natural Language Processing}, pages
  523--530, Morristown, NJ, USA, 2005. Association for Computational
  Linguistics.

\bibitem[MTF04]{Mihalcea2004}
Rada Mihalcea, Paul Tarau, and Elizabeth Figa.
\newblock Pagerank on semantic networks, with application to word sense
  disambiguation.
\newblock In {\em COLING '04: Proceedings of the 20th international conference
  on Computational Linguistics}, Morristown, NJ, USA, 2004. Association for
  Computational Linguistics.

\bibitem[PD09]{Poon2009}
Hoifung Poon and Pedro Domingos.
\newblock Unsupervised semantic parsing.
\newblock In {\em Proceedings of the 2009 Conference on Empirical Methods in
  Natural Language Processing}, pages 1--10, Singapore, August 2009.
  Association for Computational Linguistics.

\bibitem[Pro13]{HoTT2013}
The Univalent~Foundations Program.
\newblock {\em Homotopy Type Theory: Univalent Foundations of Mathematics}.
\newblock Institute for Advanced Study, 2013.

\bibitem[RVG05]{RelEx}
Mike Ross, Linas Vepstas, and Ben Goertzel.
\newblock Relex semantic relationship extractor.
\newblock http://opencog.org/wiki/RelEx, 2005.

\bibitem[SM07]{Mihalcea2007}
Ravi Sinha and Rada Mihalcea.
\newblock Unsupervised graph-basedword sense disambiguation using measures of
  word semantic similarity.
\newblock In {\em ICSC '07: Proceedings of the International Conference on
  Semantic Computing}, pages 363--369, Washington, DC, USA, 2007. IEEE Computer
  Society.

\bibitem[ST91]{Sleator1991}
Daniel Sleator and Davy Temperley.
\newblock Parsing english with a link grammar.
\newblock Technical report, Carnegie Mellon University Computer Science
  technical report CMU-CS-91-196, 1991.

\bibitem[ST93]{Sleator1993}
Daniel~D. Sleator and Davy Temperley.
\newblock Parsing english with a link grammar.
\newblock In {\em Proc. Third International Workshop on Parsing Technologies},
  pages 277--292, 1993.

\bibitem[Ste90]{Steele1990}
James Steele, editor.
\newblock {\em Meaning-Text Theory: Linguistics, Lexicography, and
  Implications}. University of Ottowa Press, 1990.

\bibitem[Tem07]{Temper2007}
David Temperley.
\newblock Minimization of dependency length in written english.
\newblock {\em Cognition}, 105:300--333, 2007.

\bibitem[Tes59]{Tesn1959}
Lucien Tesni\`{e}re.
\newblock {\em \'{E}l\'{e}ments de syntaxe structurale}.
\newblock Klincksieck, Paris, 1959.

\bibitem[WP-a]{WP-Argument}
Argument.
\newblock $http://en.wikipedia.org/wiki/Arguments_(linguistics)$.

\bibitem[WP-b]{WP-SAT}
Boolean satisfiability problem.
\newblock $http://en.wikipedia.org/wiki/Boolean_satisfiability_problem$.

\bibitem[WP-c]{WP-CMI}
Conditional mutual information.
\newblock $http://en.wikipedia.org/wiki/Conditional_mutual_information$.

\bibitem[WP-d]{WP-DPLL}
Dpll algorithm.
\newblock $http://en.wikipedia.org/wiki/DPLL_algorithm$.

\bibitem[WP-e]{WP-Predicate}
Predicate.
\newblock $http://en.wikipedia.org/wiki/Predicate_(grammar)$.

\bibitem[Yur98]{Yuret1998}
Deniz Yuret.
\newblock {\em Discovery of Linguistic Relations Using Lexical Attraction}.
\newblock PhD thesis, MIT, 1998.

\end{thebibliography}

\appendix

\section*{Appendix A: Meaning-Text Theory}

The most comprehensive theory of meaning, and its conversion to text
is Meaning-Text Theory (MTT) \cite{Mel'cuk1987,Kahane2003,Milicevic2006,Steele1990}.
Although the theory itself is primarily oriented towards the generation
of text from meaning, I believe that its representation of meaning
is ideal for extracting meaning from text. MTT is not only compatible
with dependency grammar, it can be thought of as an extension to it;
it provides rules for converting meaning into dependency graphs. These
rules are quite specific, and thus lend themselves to an algorithmic
implementation: this is another strength of MTT. Possibly the most
important contribution of MTT to linguistics is the discovery of ``lexical
functions'', which map concepts to words. 

\begin{figure}[bh]
\caption{A Semantic Representation\label{fig:A-SemR-representation}}

\begin{centering}
\includegraphics[width=0.7\columnwidth]{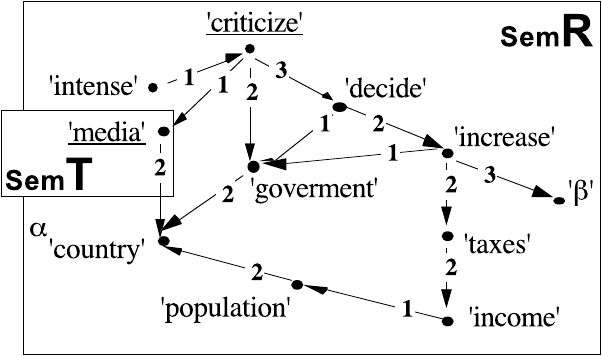}
\par\end{centering}

An example of a semantic representation (SemR) from Meaning-Text Theory.
This network of predicate-argument arrows captures the meaning of
the sentence ``\emph{The media harshly criticized the Government
for its decision to increase income taxes.}`` Here, ``\emph{media}''
is the topic or theme, and what the media is saying is the rheme.
Compare this to figure \ref{fig:Alternative-Sematic-Topic}.

Figure taken from \cite{Milicevic2006}

\rule[0.5ex]{1\columnwidth}{1pt}
\end{figure}

At it's core, MTT captures meaning with a ``semantic representation''
(SemR). A semantic representation is a network of predicate-argument
relations (in the sense of a linguistic predicate\cite{WP-Predicate}
and linguistic argument\cite{WP-Argument}). An example of such a
network is shown in figure \ref{fig:A-SemR-representation}. Each
arrow in the figure is a semantic dependency, that is, a predicate-argument
relationship (arguments are also called '\emph{semantic actants}'
in the theory). The arrows are labeled with numbers corresponding
to the valency or number of arguments that a node may have. Thus,
for example, the verb ``\emph{criticize}'' has a valency of three:
``\emph{X criticizes Y for Z}''. Nodes in the graph may be primitive
or atomic, in which case they are called '\emph{semes}', although
they may also have structure and are thus called '\emph{semantemes}'.
Note that semantemes are highly lexicalized: that is, the definition
of a semanteme specifies the number of actants, and their roles. In
essence, semantemes correspond to dictionary entries (a '\emph{lexis}'
is a dictionary).

\begin{figure}
\begin{centering}
\caption{Alternative Semantic Topic\label{fig:Alternative-Sematic-Topic}}

\par\end{centering}

\begin{centering}
\includegraphics[width=0.7\columnwidth]{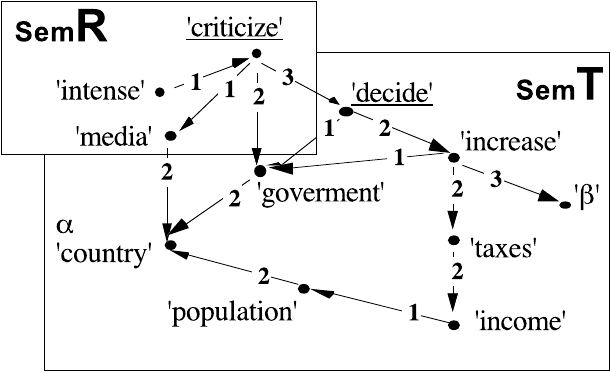}
\par\end{centering}

An alternative partitioning of a semantic network into theme and rheme.
In this case, the topic is ``\emph{the Governments decision to raise
taxes}'', and what is being said about this topic is that the media
is sharply critical of it. In words, one could say that ``\emph{The
Government\textquoteright{}s decision to increase income taxes was
severely criticized by the media}.'' Note that this sentence is more-or-less
synonymous with that given in figure \ref{fig:A-SemR-representation}.

Figure taken from \cite{Milicevic2006}

\rule[0.5ex]{1\columnwidth}{1pt}
\end{figure}

More properly, the network described above is a 'SemS' (semantic structure);
it is but one part, although a core part, of a SemR. The network captures
the propositional/situational meaning at the core. The other three
parts of SemR are termed Sem-CommS, RhetS and RefS. The Sem-CommS
(sematic-communicative structure) captures the communicative intent;
the '\emph{theme}' (what is being talked about) and the '\emph{rheme}'
(what is being said about the theme). For example, in figure \ref{fig:A-SemR-representation},
the topic is 'media', and the rheme is what the media is talking about
(a raise in taxes). There is no unique assignment of themes and rhemes
to a network, thus, for example, in figure \ref{fig:Alternative-Sematic-Topic},
the theme is the raise in taxes, and the rheme is what is being said
about it: the media is criticizing it. The Sem-CommS also has several
other parts, the most important of which is distinguishing what is
new information from what is given; that is, differentiating what
is asserted from background pre-suppositions. A general axiom of MTT
is that meaning is invariant under paraphrasing: thus, for example,
``\emph{The media harshly criticized the Government for its decision
to increase income taxes}`` and ``\emph{The Government\textquoteright{}s
decision to increase income taxes was severely criticized by the media}''
are roughly synonymous; thus the network underlying the two figures
\ref{fig:A-SemR-representation} and \ref{fig:Alternative-Sematic-Topic}
is the same. Tightening down the distinction between new and pre-supposed
information breaks synonymy; it narrows the range of sentences that
can be considered synonymous, of saying the same thing. Thus, MTT
can capture the finer points of a speech-writer's art: the careful
crafting of sentences to convey a very specific meaning.

The other parts of SemR are RhetS and RefS. RhetS specifies the rhetorical
style used in converting a semantic network to text (such as headline-news,
where sentences are clipped ('\emph{Thieves rob Bank}'); informal
speech with lots of lulz, typoes, ikr, smh and smiley winks ;->, or
proper newspaper-English.) The RefS captures the referential structure:
the references to concrete objects in the (model of the) external
environment ('\emph{the ball rolled under the sofa}' refers to a specific
ball and sofa) or anaphora (the pronoun '\emph{she}' in '\emph{she
waved goodbye}' refers to a specific person).

MTT distinguishes between several different forms or levels of representation;
the above described is SemR, the semantic representation. There is
also a SyntR, the syntactic representation, roughly corresponding
to a dependency parse, as well as MorphR, a morphological representation,
where not only has the word-order been chosen, but so have verb tenses,
inflections of nouns and the like. A PhonR representation gives the
spoken, phonemic form of a sentence. Each of these levels has additional
structures that capture important information (such as verb tense,
choice of adjective or adverb modifiers and the like). Between each
of these levels are a set of correspondence rules that translate structures
at one level to those in another. Roughly speaking, correspondence
rules can be thought of as functors that map networks at one level
to those of another. MTT attempts to treat these rules as mechanically
as possible, with an eye to algorithmic implementation. 

Perhaps the most important contribution of MTT to linguistics is the
discovery of '\emph{lexical functions}' (LF's); these appear in the
correspondence rules. Lexical functions bind a meaning to a lexeme.
For example, the LF \noun{Magn()} is a function that specifies a list
of appropriate words for expressing magnitude. One then has \noun{Magn}(\emph{rain})
== \emph{torrential}|\emph{hard}, \noun{Magn}(\emph{wind}) == \emph{strong}
and \noun{Magn}(\emph{emotion}) == \emph{hot.} The point here is that
in each case, a magnitude is being expressed; yet, it is context-specific:
one would not normally say '\emph{hot rain}', '\emph{hard wind}' or
'\emph{torrential emotion}'; the LF specifies the allowable modifiers
for a given noun. The LF \noun{Magn()} is broad in scope, applying
to nouns in general; not all LF's are universal in this way. Some
can have a very narrow scope: for example, 'leap' only applies to
'year'. Another example is the subject LF \noun{S\textsubscript{1}()
}that indicates authorship: \noun{S\textsubscript{1}(}\emph{crime}\noun{)}
== \emph{perpetrator},\noun{ S\textsubscript{1}(}\emph{book}\noun{)
}== \emph{author}. A third example is the quasi-subject function that
gives noun-equivalents for verbs: \noun{QS}\textsubscript{0}(\emph{criticize})
== \emph{criticism}. MTT defines many different lexical functions;
a sampling of these for the verb '\emph{criticize}', taken from \cite{Milicevic2006},
are:
\begin{lyxlist}{00.00.0000}
\item [{~}]~

\begin{lyxlist}{00.00.0000}
\item [{\noun{QSyn}:}] attack, disapprove, reproach 
\item [{\noun{QAnti}:}] praise, congratulate
\item [{\noun{QS}\textsubscript{0}:}] criticism
\item [{\noun{S}\textsubscript{1}:}] critic
\item [{\noun{A}\textsubscript{1}:}] critical {[}of N{]}
\item [{\noun{Magn}:}] bitterly, harshly, seriously, severely, strongly
\item [{\noun{Magn}\textsubscript{Quant}:}] all the time, relentlessly,
without stopping
\item [{\noun{AntiMagn}:}] half-heartedly, mildly
\item [{\noun{AntiVer}:}] unjustly, without reason 
\end{lyxlist}
\end{lyxlist}
\,

As should be apparent from the above, lexical functions provide all
of the relations that can be found in resources like WordNet (\emph{viz,}
a machine-readable catalog of word-senses, their synonyms, antonyms
and hypernyms), while providing a more comprehensive view of the nature
of these relations. Similarly, the predicate-argument lexical entries
of MTT resemble the frames of FrameNet. Unlike FrameNet, however,
MTT describes the mechanisms by which these semantic frames become
connected to syntactic representations. It provides a more comprehensive
description of how frames interact with other aspects of grammar.

To recap: meaning is captured by referential structures, obtained
from semantic structures built of lexical functions. The goal of learning
language is to ascend through a hierarchy of structures, from raw
text, through dependency grammars, up to referential disambiguation
of a world-model. To make sure that learning takes place at a reasonable
pace, deep-learning style reinforcement must happen at each layer,
so that the simpler, shallower layers (the syntactic layers) are somewhat
developed before the semantic layers are attacked; yet that the deeper
layers can also guide correct learning at the shallower layers. MTT,
as opposed to some other theory or framework, seems appropriate for
providing a basic framework of what must be learned. This is in part
because MTT seems to be the most comprehensive theory describing not
only the various layers, but also algorithmic mechanisms of transforming
one layer into another. 

The language learning system must learn lexical functions on its own;
it must learn how to pick out predicate-argument structures; these
are not pre-supposed. Rather, MTT provides a viewpoint by which the
success of the learning system might be judged: instead of treating
learning as a black-box, one might expect being able to examine what
is being learned, and one might reasonably expect it to resemble the
outlines of MTT.

\section*{Appendix B: Mutual Information}

This appendix provides some gymnastics for working with probabilities
associated with structures and relations. It is provided only because
such discussions are rare in the literature. Only the very simplest
cases is worked here: the mutual information between a pair of words.
Hopefully, generalizations then become obvious.

Let $P(R(w_{l},w_{r}))$ represent the probability (frequency) of
observing two words, $w_{l}$ and $w_{r}$ in some relationship or
pattern $R$. Typically, in Link Grammar, it would be a linkage of
type $t$ connecting word $w_{l}$ on the left to word $w_{r}$ on
the right; however, the relation $R$ can be more general than that.

The simplest model has only one type $t$, the ANY type, and assigns
equal probabilities to all words. But we know all words are not equi-probable,
so let $P(w)$ be the probability of observing word $w$. We know
from experience this is a Zipfian distribution. We are then interested
in the conditional probability $P(R(w_{l},w_{r})|w_{l},w_{r})$ of
observing the two words $w_{l}$ and $w_{r}$ in a relation $R=R(w_{l},w_{r})$,
given that the two individual words were observed. From the definition
of conditional probabilities, one has that 
\[
P(R)=P(R|w_{l},w_{r})P(w_{l})P(w_{r})
\]
or, equivalently, that 
\[
P(R|w_{l},w_{r})=\frac{P(R)}{P(w_{l})P(w_{r})}
\]
Here, the relation $R$ encompasses several facts: that one word is
to the left of the other, and that they are connected by a certain
link-type, as well as capturing other 'ambient' information, perhaps
such as other nearby words.

It is important here to harmonize this with the notation used by Yuret\cite{Yuret1998}
and commonly employed in the MST parser literature. There, a probability
$P(w_{l},w_{r})$ is defined of seeing the ordered pair; that is,
the relation $R$ is implicit. To make it explicit, we should write:
$P(w_{l},w_{r})=P(R(w_{l},w_{r}))$ to indicate the relation explicitly,
and to note that the order of the positions in the relation matter.
Yuret also uses the notation $P(w_{l},*)$ and $P(*,w_{r})$ for wild-card
summations, defined as 
\[
P(w_{l},*)=\sum_{w_{r}}P(w_{l},w_{r})\qquad\mbox{and}\qquad P(*,w_{r})=\sum_{w_{l}}P(w_{l},w_{r})
\]
In practical use, one quickly observes that $P(w_{l},*)$ is almost
equal to $P(w_{l})$ but not quite, since $P(w_{l},*)$ is the probability
of seeing $w_{l}$ within the certain relationship or pattern, which
must be less than the prior probability of observing $w_{l}$ in general.
Thus, one has $P(w_{l},*)\le P(w_{l})$ which can be viewed as a conditional
probability:
\[
P(w_{l},*)=P(R(w_{l},*))=P(R(w_{l},*)|w_{l})P(w_{l})
\]
In practice, then, for word-pairs, one has that $P(R|w_{l})$ is almost
equal to 1, but not quite. Inserting this into the above gives 
\[
P(R(w_{l},w_{r})|w_{l},w_{r})=\frac{P(R(w_{l},w_{r}))\ P(R(w_{l},*)|w_{l})\ P(R(*,w_{r})|w_{r})}{P(w_{l},*)P(*,w_{r})}
\]
Re-ordering this gives
\begin{equation}
\frac{P(R(w_{l},w_{r})|w_{l},w_{r})}{P(R(w_{l},*)|w_{l})\ P(R(*,w_{r})|w_{r})}=\frac{P(w_{l},w_{r})}{P(w_{l},*)P(*,w_{r})}\label{eq:basic pair}
\end{equation}
The right hand side above is recognizable from Yuret's work; he defines
the mutual information as 
\[
\mbox{MI}(w_{l},w_{r})=\log_{2}\,\frac{P(w_{l},w_{r})}{P(w_{l},*)P(*,w_{r})}
\]
so that large positive MI is associated with words that occur together
only with themselves (\emph{e.g.} \emph{``Northern Ireland}'', from
his examples.) So, on the right, we have that $P(w_{l},w_{r})$ is
usually very small, and that $P(w,*)\approx P(*,w)\approx P(w)$ subject
to the inequality given before. 

The LHS of equation (1) shows how to properly
normalize conditional probabilities for general structures when performing
``frequent subgraph mining''. First, we have observationally seen
that $P(R(w_{l},*)|w_{l})\approx P(R(*,w_{r})|w_{r})\approx1$, and
thus must conclude that $P(R(w_{l},w_{r})|w_{l},w_{r})$ is 'large';
much larger than (unconditional) word-pair frequencies. 

The LHS of equation (1) then demonstrates how
to obtain conditional entropies in general. Thus, given an $n$-point
relation $R(x_{1},x_{2,}\cdots,x_{n})$ one computes first the unconditional
probability $P(R(x_{1},x_{2,}\cdots,x_{n}))$. The conditional probability
is then obtained as usual:
\[
P(R(x_{1},x_{2,}\cdots,x_{n})\,|\, x_{1},x_{2,}\cdots,x_{n})=\frac{P(R(x_{1},x_{2,}\cdots,x_{n}))}{P(x_{1})P(x_{2})\cdots P(x_{n})}
\]
The entropy is then build recursively by normalizing by the probability
of wild-card relations:
\[
MI(R(x_{1},x_{2,}\cdots,x_{n})\,|\, x_{1},x_{2,}\cdots,x_{n}))=\log_{2}\frac{P(R(x_{1},x_{2,}\cdots,x_{n})\,|\, x_{1},x_{2,}\cdots,x_{n}))}{P(R(*,x_{2,}\cdots,x_{n})\,|\, x_{2,}\cdots,x_{n}))\, P(R(x_{1},*,\cdots,x_{n})\,|\, x_{1},x_{3,}\cdots,x_{n}))\cdots}
\]

The point of this derivation is to provide a simpler practical formulation
for working with structural relations in language. Most presentations
of conditional mutual information obscure the structural relationships,
by hiding the wild-card summations in a different notation that makes
it hard to discern their presence; see, for example \cite{WP-CMI}
for a demonstration of an equivalent but more opaque notation. Do
observe that the RHS above is a special case, though, where the $x_{k}$
appearing in the relation are identical to those appearing in the
condition. When these are not the same, then a summation is required,
as usual:
\[
I(R(X_{1},X_{2,}\cdots,X_{n})\,|\, Z))=\sum_{z\in Z}P(z)\sum_{x_{k}\in X_{k}}P(R(x_{1},x_{2,}\cdots,x_{n})\,|\, Z))\log_{2}\frac{P(R(x_{1},x_{2,}\cdots,x_{n})\,|\, Z))}{P(R(*,x_{2,}\cdots,x_{n})\,|\, Z))\cdots}
\]
The difference between this and the previous equation is that, when
the $X_{k}=\{x_{k}\}$ are all singleton sets, and $Z=\{z=(x_{1},x_{2},\cdots,x_{n})\}$
is likewise a singleton set, then the summations disappear. One is
left with a mutual information, scaled by the (unconditioned) probability
of seeing the particular pattern. Because the pattern may in fact
be very rare, this is not as useful in practical experimentation than
the renormalized mutual information.
\end{document}